\documentclass[letterpaper]{article} % DO NOT CHANGE THIS
\usepackage{aaai24}  % DO NOT CHANGE THIS
\usepackage{times}  % DO NOT CHANGE THIS
\usepackage{helvet}  % DO NOT CHANGE THIS
\usepackage{courier}  % DO NOT CHANGE THIS
\usepackage[hyphens]{url}  % DO NOT CHANGE THIS
\usepackage[pdftex]{graphicx} % DO NOT CHANGE THIS
\usepackage{natbib}  % DO NOT CHANGE THIS AND DO NOT ADD ANY OPTIONS TO IT
\usepackage{caption} % DO NOT CHANGE THIS AND DO NOT ADD ANY OPTIONS TO IT
\usepackage{algorithm}
\usepackage{algorithmic}

\usepackage{newfloat}
\usepackage{listings}
\DeclareCaptionStyle{ruled}{labelfont=normalfont,labelsep=colon,strut=off} % DO NOT CHANGE THIS
\floatstyle{ruled}
\newfloat{listing}{tb}{lst}{}
\floatname{listing}{Listing}
\usepackage{subcaption}
\usepackage{multirow}
\usepackage{amssymb}
\usepackage{booktabs}
\usepackage{xcolor}
\usepackage{pifont}
\usepackage{colortbl}
\usepackage{bm}
\usepackage{varwidth}
\usepackage{nicematrix}

\definecolor{darkergreen}{RGB}{21, 152, 56}
\definecolor{red2}{RGB}{252, 54, 65}
\definecolor{gray}{gray}{0.6}
\definecolor{ballblue}{rgb}{0.13, 0.67, 0.8}

\newcommand\jh[1]{\textcolor{black}{#1}}

\newcommand\jyyoo[1]{\textcolor{black}{#1}}
\newcommand\hsjeong[1]{\textcolor{black}{#1}}
\newcommand\hskim[1]{\textcolor{black}{#1}}
\newcommand\jhaaai[1]{\textcolor{black}{#1}}
\newcommand\cam[1]{\textcolor{black}{#1}}

\title{ProxyDet: Synthesizing Proxy Novel Classes via Classwise Mixup \\for Open-Vocabulary Object Detection}
\author{
    Joonhyun Jeong\textsuperscript{\rm 1,\rm 2},
    Geondo Park\textsuperscript{\rm 2},
    Jayeon Yoo\textsuperscript{\rm 3},
    Hyungsik Jung\textsuperscript{\rm 1},
    Heesu Kim\textsuperscript{\rm 1}\footnote{Corresponding Author.}
}
\affiliations{
    \textsuperscript{\rm 1}NAVER Cloud, ImageVision\\
    \textsuperscript{\rm 2}Korea Advanced Institute of Science and Technology (KAIST) \\
    \textsuperscript{\rm 3}Seoul National University \\
    \{joonhyun.jeong, hyungsik.jung, heesu.kim89\}@navercorp.com, geondopark@kaist.ac.kr, jayeon.yoo@snu.ac.kr
}

\usepackage{bibentry}
\begin{document}

\maketitle

%%%%%%%%% ABSTRACT
\begin{abstract}
     Open-vocabulary object detection (OVOD) aims to recognize \jhaaai{novel} objects whose categories are not included in \cam{the} training set.
     In order to classify \jhaaai{these} unseen classes during training, many OVOD frameworks leverage the zero-shot capability of largely pretrained vision and language models, such as CLIP.
     \jhaaai{To further improve generalization on the unseen novel classes, several approaches proposed to additionally train with pseudo region labeling on the external data sources that contain a substantial number of novel category labels beyond the existing training data.}
     \jhaaai{Albeit its simplicity, these pseudo-labeling methods still exhibit limited improvement with regard to the \cam{truly unseen} novel classes that were not pseudo-labeled.}
     \jhaaai{In this paper, we present a novel, yet simple technique that helps generalization on the overall distribution of novel classes.}
     \jhaaai{Inspired by our observation that numerous novel classes reside within the convex hull constructed by the base (seen) classes in the CLIP embedding space, we propose to synthesize proxy-novel classes approximating novel classes via linear mixup between a pair of base classes. By training our detector with these synthetic proxy-novel classes, we effectively explore the embedding space of novel classes. The experimental results on various OVOD benchmarks such as LVIS and COCO demonstrate superior performance on novel classes compared to the other state-of-the-art methods.} \cam{Code is available at \url{https://github.com/clovaai/ProxyDet}.}

\end{abstract}

%%%%%%%%% BODY TEXT
\section{Introduction}
\label{sec:intro}

% 티저 - Mixup 그림 (첫페이지)
%\input{figures/teaser}

% 두번째 페이지 (첫페이지의 근거: mixup embedding의 novel class와의 similarity distribution 비교.)
% \input{figures/table1_figure2_mixed}

% 1문단. detection task에 대한 소개. open-vocab detection이 도입되게된 배경? 
% object detection task에 대한 소개.

 Object detection is a task in which the model simultaneously predicts where all objects are located within an image and to which class each object belongs.
 Conventional \jhaaai{closed-set} object detection frameworks~\citep{retinanet, mask_rcnn, lin2017feature} have limited the range of discernible objects to a fixed set of base classes, which are usually the classes annotated in training sets.
 Recently, \cam{the} performance of those frameworks on the fixed set of classes has largely evolved, building upon the success of vision transformer architectures \cite{dosovitskiy2020image, liu2022swin, carion2020end, zong2022detrs, chen2022group} and large-scale pretraining techniques \cite{su2022towards, chen2022group, wang2022image}. 
 However, beyond improving the performance on the fixed class set, \emph{open-vocabulary object detection} (OVOD) frameworks \cite{vild, ovrcnn} have emerged to tackle the limited range of discernible objects. 
 The objective is to detect a set of novel classes which are not seen \jhaaai{during} training, without newly labeling the bounding boxes and class labels of these novel classes which is labor-intensive and non-scalable.
In order to achieve open-vocabulary property, OVOD frameworks should expand \jhaaai{their classifier} for novel classes. % \jh{\sout{with only the names of the classes at inference.}}
%without explicit access to \jh{visual information of novel classes.} % visual \hskim{\sout{and textual}} information of novel classes.
Therefore, previous OVOD frameworks~\cite{vild, detic, f-vlm} adopt pretrained vision-language model (VLM), \jhaaai{CLIP}~\cite{clip}, to obtain the \jhaaai{weights} of their classifier by prompting the names of base and novel classes.
These VLM models provide an interface (i.e., encoder) that maps textual information (e.g., class name) to an embedding space where embeddings of semantically matched images are closely located.
% These VLMs provide encoders that map arbitrary images and text prompts to an embedding space where embeddings of semantically matched images or text have high cosine similarity. 
In line with previous work, our study also adopts this approach to enable open-vocabulary detection.

\jhaaai{In order to further explicitly expand the range of detection vocabulary on the novel classes}, recent OVOD frameworks~\cite{detic, promptdet, vl-plm, pb_ovd, vldet} rely on \jhaaai{pseudo-labeling a subset of novel classes given in} the inexpensive supplementary datasets~\cite{ridnik2021imagenet} consisting of \jh{pairs of image and its class label (or caption) without box annotations.}
% In order to improve the classification performance on novel classes, recent OVOD frameworks~\cite{detic, promptdet, vl-plm, pb_ovd, rkd, vldet} rely on the inexpensive supplementary datasets consisting of image-level labels or pairs of image and caption, without box annotations. 
The missing box annotations are predicted using \jhaaai{pretrained} region proposal network (RPN) and their class labels are assigned using image-level labels~\cite{detic, rkd} and the words within the captions~\cite{vldet}, \jhaaai{which contain a substantial number of novel classes.}
% The missing box annotations are predicted using box predictors such as RPN.
% 위 문장대신?
% These external data improve object detection performance by pseudo-labeling object regions through some heuristics \cite{detic} or by using largely pre-trained box predictors \cite{rkd}.
These pseudo-labeling approaches are essentially identical to the conventional \jhaaai{closed-set} object detection, the only difference being that the human-labeled box annotations are not used. 
Consequently, it has limitations in fundamentally solving generalization to \jhaaai{the \cam{truly unseen} novel classes that were not pseudo-labeled}, only to improve \jhaaai{performance} for the pseudo-labeled novel classes as shown in Table \ref{table:table1_nonovl_ap}.
% Despite its efficacy in improving the classification performance on novel classes by providing external visual information, we observe that the improvement is only limited to the novel classes included in the additional datasets. 
% Consequently, these pseudo-labeling approaches are essentially identical to the conventional object detection, the only difference being that the human-annotated box annotations are not used.

% \begin{figure}[t]
%     \centering
%     % \includegraphics[width=0.95\linewidth]{figures/teaser.png}
%     % \includegraphics[width=0.95\linewidth]{figures/teaser_updated.png}
%     % \includegraphics[width=0.95\linewidth]{figures/teaser_updated_2.png}
%     \includegraphics[width=0.95\linewidth]{LaTeX/figures/teaser_updated.png}
    
%     \caption{\ctv{\textbf{Motivation of ProxyDet.} Linear combination of known (base) class embeddings in CLIP \cite{clip} embedding space generates proxy-novel representations. Learning these synthesized representations equip OVOD models to well-generalize on unknown novel classes and consequently increase the vocabulary set.}}
%     \vspace{-7mm}
%     \label{fig:teaser_mixup_concept}
% \end{figure}

\begin{figure}[t]
    \centering
    \begin{subfigure}[b]{0.475\textwidth}
        \centering
        % \hspace{-7mm}
        % \includegraphics[width=\textwidth]{figures/tsne4.png}
        \includegraphics[width=\textwidth]{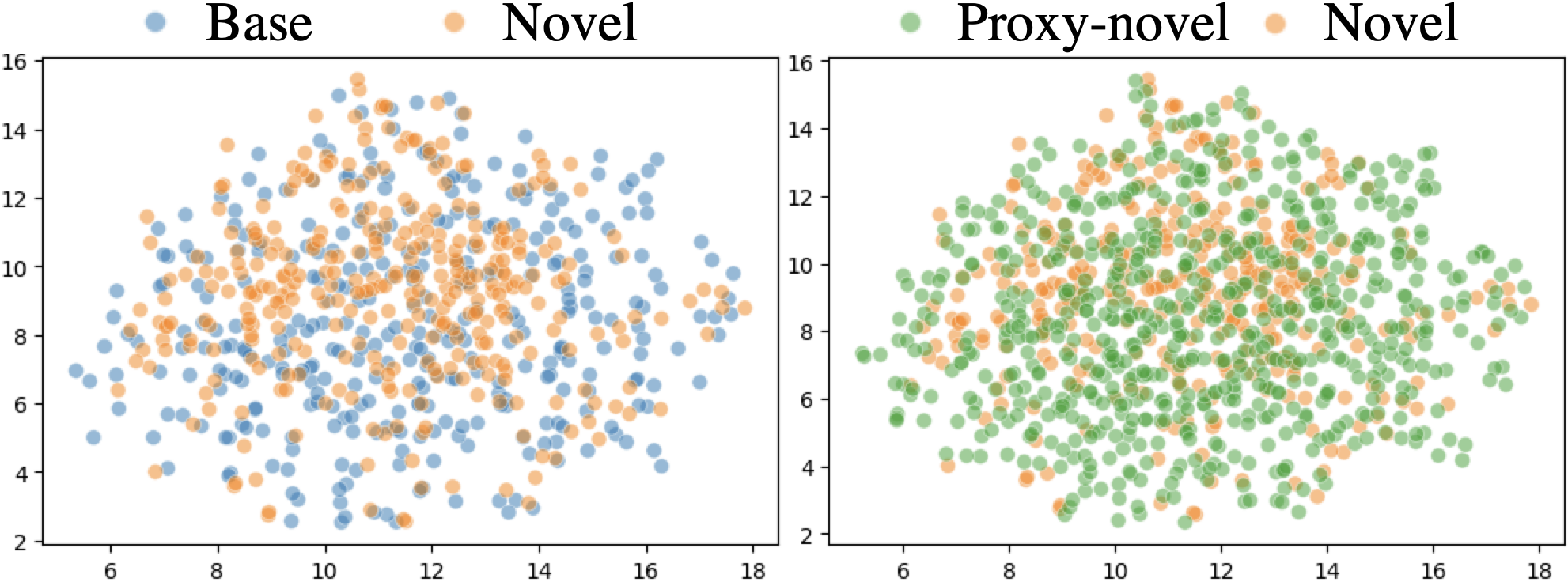}
        \caption{\cam{UMAP~\cite{mcinnes2018umap} visualization of CLIP text embeddings for base/novel classes (left), proxy-novel/novel classes (right) in LVIS dataset.}}
        \label{fig:tsne}
        \vspace{1mm}
    \end{subfigure}
    \begin{subfigure}[b]{0.4\textwidth}
        \centering
        \includegraphics[width=\textwidth]{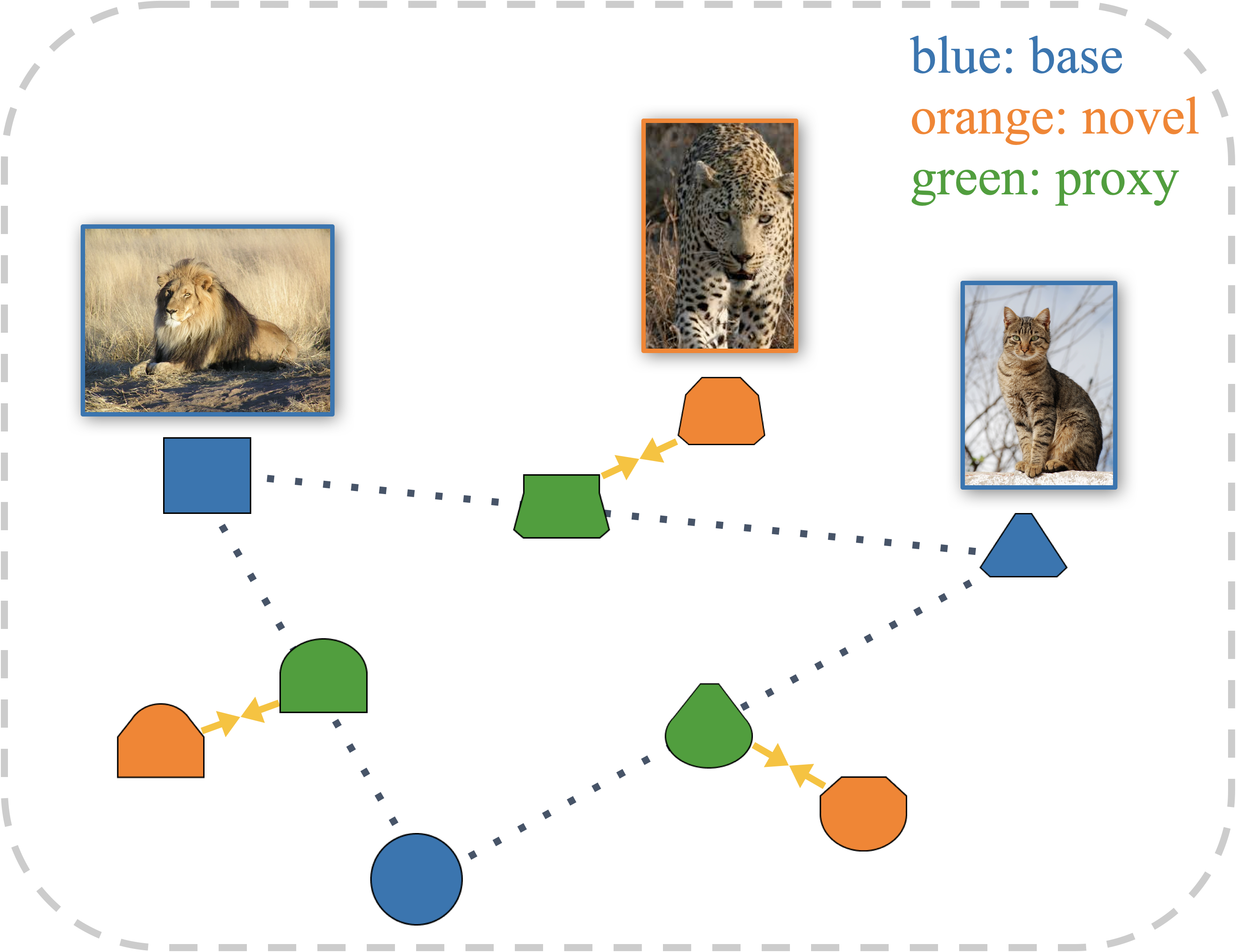}
        \caption{Proxy-novel class synthesis via convex combination of the base classes in CLIP embedding space.}
        \label{fig:concept}
    \end{subfigure}
    % \hspace{10mm}
    \caption{Visualization of embedding space for all class groups (a) and our proposed proxy-novel synthesis (b).}
    \label{fig:teaser}
    \vspace{-5.5mm}
\end{figure}

\begin{figure}[t]
    \centering
        \centering
        \includegraphics[width=0.31\textwidth]{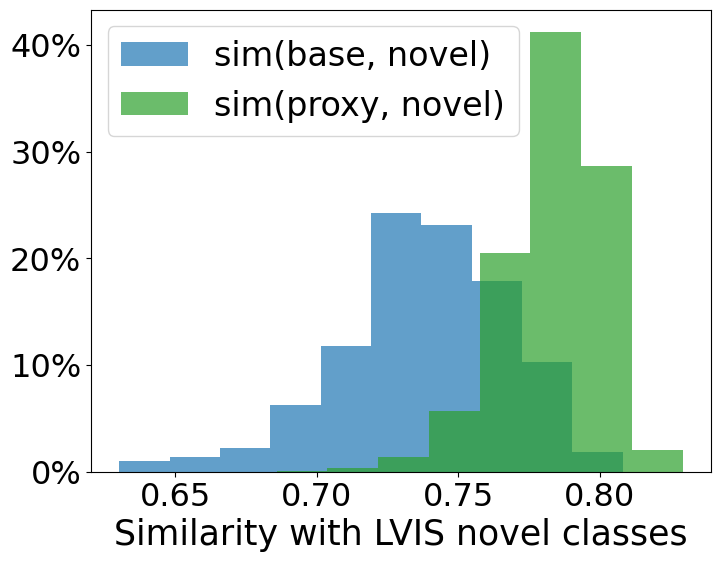}
        \vspace{-3mm}
        \caption{\jhaaai{\cam{Histogram of similarity} between training class groups (base and proxy-novel) and novel classes.} 
        We compute pairwise cosine similarity between all the classes in each of training class group and novel classes in LVIS dataset, using text representations obtained from CLIP.} 
        \label{fig:novel_similarity_distribution}
    \label{fig:novel_similarity}
    \vspace{-6mm}
\end{figure}

\jhaaai{To overcome this limitation, we present ProxyDet, a new OVOD framework that helps better generalization on the overall distribution of novel classes including these \cam{truly unseen novel} ones that have never been presented during training. Empirically, we observe that a wide range of novel classes resides within the convex hull constructed by the base classes in the VLM embedding space \cam{as shown in the left side of} Figure \ref{fig:tsne}. Motivated by this observation, we propose a synthesis of \textit{proxy-novel} classes, which approximates novel classes via convex combination between the base classes (Figure~\ref{fig:concept}). Notably, we observe that the overall distribution of novel classes can be better approximated through our proposed proxy-novel synthesis as shown in Figure~\ref{fig:novel_similarity}. Learning on these proxy-novel classes provides an extensive proximal exploration of numerous novel classes during training (\cam{the right side of Figure~\ref{fig:tsne}}), thus effectively expanding the detection vocabulary. Consequently, as shown in Table~\ref{table:table1_nonovl_ap}, our method improves Average Precision (AP) on the \cam{non-pseudo-labeled} novel classes up to \cam{$3.4$}, while the previous pseudo-labeling approach~\cite{detic} rather degrades by \cam{$0.3$} (See the Appendix\footnote{\cam{The full extended paper including the appendix is available at \url{https://arxiv.org/abs/2312.07266}.}} for the performance of the other pseudo-labeling methods).
% the similarities to the novel classes are higher for the proxy-novel classes compared to base classes.
% These proxy-novel classes provide an extensive exploration of numerous novel classes during training, thus effectively expanding the detection vocabulary.  
% Consequently, the proxy-novel classes provide the object detector with the knowledge helping to understand novel classes, which has not been obtained from base classes.
}

\begin{table}[t]
% \vspace{-1mm}
\centering
\smallskip\noindent
\tabcolsep=0.08cm

\vspace{-1mm}
\resizebox{0.965\linewidth}{!}{
\begin{NiceTabular}{c|ccc}
\toprule
\multirow{2}{*}{\begin{tabular}[c]{@{}c@{}} Method \end{tabular}} & 
\multicolumn{3}{c}{$\text{AP}_r$}\\
\cmidrule(lr){2-4}
& Non-pseudo & \textcolor{black}{Pseudo} & Overall  \\
        
\midrule

\multirow{2}{*}{\begin{tabular}[c]{@{}c@{}}  Pseudo-labeling~\cite{detic}  \end{tabular}} & 
\multirow{2}{*}{\begin{tabular}[c]{@{}c@{}} 19.4 \\ \textcolor{black}{(-0.3)} \end{tabular}} &
\multirow{2}{*}{\begin{tabular}[c]{@{}c@{}} \textcolor{black}{25.9} \\ \textcolor{black}{(+10.0)}
\end{tabular}} &
\multirow{2}{*}{\begin{tabular}[c]{@{}c@{}} \textcolor{black}{24.6} \\ \textcolor{black}{(+7.9)} \end{tabular}} \\ 
& & &   \\

\multirow{2}{*}{\begin{tabular}[c]{@{}c@{}} Ours  \end{tabular}} & 
\multirow{2}{*}{\begin{tabular}[c]{@{}c@{}} 23.1 \\ \textbf{\textcolor{black}{(+3.4)}} \end{tabular}} &
\multirow{2}{*}{\begin{tabular}[c]{@{}c@{}} \textcolor{black}{27.0} \\ \textcolor{black}{(+11.1)}
\end{tabular}} &
\multirow{2}{*}{\begin{tabular}[c]{@{}c@{}} 26.2 \\ \textcolor{black}{(+9.5)} \end{tabular}} \\ 
& & &   \\
\bottomrule

\end{NiceTabular}
}

\vspace{-2mm}
\caption{\jhaaai{Effectiveness of pseudo-labeling and our method on the novel classes. The evaluation is conducted on novel classes that are separated based on whether they are pseudo-labeled or not, abbreviated as "Non-pseudo" and "Pseudo". "Overall" denotes performance on the overall novel classes.}
% \cam{The quantities in the parentheses denote the amount of performance change compared to the baseline, VILD~\cite{vild}} 
\cam{The quantities in the parentheses indicate the amount of performance changes relative to the vanilla case without applying pseudo-labeling or our method.}
%without applying pseudo-labeling or ours.}
}

\vspace{-3.7mm}
\label{table:table1_nonovl_ap}
\end{table}

% 5문단. 실험 결과 내용.
Our extensive experiments on various OVOD benchmarks show that this straightforward technique significantly improves classification capability for novel classes, outperforming other state-of-the-art OVOD methods. On LVIS \cite{gupta2019lvis} benchmark, our method shows largely improved performance with regard to novel class, +$1.6\ \text{AP}_r$, compared to the baseline method~\cite{detic}, while enhancing by +$2.6\ \text{AP}_{\text{novel}}$ on COCO~\cite{coco}. Furthermore, our proposed method shows the state-of-the-art performance on cross-dataset evaluation ($57.0\ \text{AP}$ on COCO and $19.1\ \text{AP}$ on Objects365), showing the superiority of our method on expanding detection vocabulary.
In summary, our contributions \cam{are} outlined as follows:

\begin{itemize}
    \item We empirically discover that mixing the representations of base classes generates proxy-novel classes that can approximate novel classes. %, which approximately facilitates novel class-aware training.

    \item Building upon the empirical observation, we propose a novel approach, introducing proxy-novel classes in training objective. By \cam{training} with \cam{the} proxy-novel classes, our proposed method encourages a model to explore the proximal representation space of novel classes.
    
    %\item Based on the observation, we propose a novel approach, classwise mixup technique, which generates pseudo proxy-novel class representations for more robust generalization of unknown novel classes and largely expanding the vocabulary set.

 %for more robust generalization of unknown novel classes and largely expanding the vocabulary set.
 % Novel class generalization without any prior information:
% 기존의 novel class에 대한 prior information을 활용하는 literature들(* Related works참고)과 달리 novel class에 대한 아무런 prior 정보 (i.e. novel class label set) 없이도 novel class representation을 학습할 수 있고, 결과적으로 novel class에 대해 더 좋은 classification 성능을 보유할 수 있다.
    
% \item We present a simple but effective add-on technique without any complex procedures.
% Simple add-on technique without offline process:
% ViLD Distillation, novel class pseudo labeling, prompt learning 등의 과정은 많은 offline 전처리 과정이 필요하므로 seemless하게 학습이 불가능하다. 그러나, ours 방법은 이러한 offline 과정 없이 단순히 network head + loss를 기존 OVD framework에 add-on 하는것만으로도 좋은 효과를 낸다
    \item Extensive experiments on various benchmarks demonstrate the superiority of our method over the other state-of-the-art OVOD methods, with a large margin of \cam{performance} improvement in detecting novel classes. % (+2.6\% in Open-Vocab COCO, +1.6\% in Open-Vocab LVIS).

% State-of-the-art Performance:
% 기존의 OVD framework이 보여주던 성능에서 아래 측면에서 더 좋은 결과를 낸다.
% W/O IN-L Setting
% W/ IN-L Setting
% (COCO experiments?)
% transfer to other datasets (COCO, Object365, etc)
\end{itemize}

% \begin{enumerate}
%     \item \todo{What is the problem you want to solve?}
%     \item \todo{Why such a problem is important?}
%     \item \todo{Is there any important impact for the community if solving such a problem? do the media discuss about its importance? it would be very nice to cite something from a news media company about this problem to make it closest to the reality of the reader/reviewer.}
%     \item \todo{how does the literature touch such a problem? what are the solutions proposed so far? what are their limitations in terms of proposed solution, datasets and difficulty of the experiments?}
%     \item \todo{how is your proposed approach supposed to defeat such limitations? what are you proposing? what is the novelty? how does it work? are the results promising? is there any difficulties in the experiments you are considering here but were not considered previously?}
% \end{enumerate}

% \todo{In summary, the contributions of this paper are}
% \begin{enumerate}
%     \item \todo{Contribution one}
%     \item \todo{Contribution two}
%     \item \todo{etc.}
% \end{enumerate}
\section{Related Work}

\paragraph{Mixup} 
Regularizing neural networks with data augmentation during training is essential to alleviate overfitting on fixed training dataset. 
Beyond the fundamental image transformations (cropping, flipping, and color jittering), image-mixing augmentations~\cite{mixup, cutmix, puzzlemix, comixup, dcutmix} have successfully resolved the overfitting problem and achieve better generalization performance on training classes, becoming a widely-accepted standard for training neural network based classifiers. 
Especially, mixup~\cite{mixup} regularize\cam{d} the overfitting on training images by presenting synthesized images made by convex combinations between existing training instances (images).
Furthermore, \citet{manifold_mixup} and \citet{geodesic_mixup} expand\cam{ed} the application of mixup to the hidden representation (embedding) space of the uni-modal and multi-modal network, respectively. While \jhaaai{these mixup variants}~\cite{mixup, cutmix, comixup, manifold_mixup, dcutmix} have primarily focused on instance-wise mixup in image space for better generalization on training classes, this paper concentrates on the class-wise mixup \cam{that explores the augmented multi-modal representation space spanned by the base training classes}, for \cam{better} understanding \cam{the} novel classes which are not presented for training.
% Built upon the multi-modal mixup \cite{geodesic_mixup}, we target to train with proxy-novel classes through mixing the class-specific representations. This technique helps to generalize on novel classes in OVOD task.
% geodesic mixup 언급 살짝
% 차별성: 1. class-wise mixup for proxy-novel class generation
% 차별성 2. 
% \vspace{-4mm}
\paragraph{Open-vocabulary object detection} 
Based on an extensive source of object detection data with high-quality annotations, object detection frameworks~\cite{retinanet, yolo, eresfd, detr} have flourished on various benchmarks~\cite{coco, pascal_voc, gupta2019lvis} where the range of classes to be detected is predefined and limited. 
In order to expand the predefined class set without the laborious annotation procedures, open-vocabulary object detection frameworks~\cite{ovrcnn, vild, regionclip, detpro, f-vlm, owlvit} have been proposed. 
The pioneering work~\cite{ovrcnn} \cam{employed} the backbone pretrained with a corpus of image-caption pairs.
% \jh{\sout{directly merge the pre-trained textual and visual uni-modal networks into the detector.}}
With the advent of massively pretrained VLMs~\cite{clip, align}, ViLD~\cite{vild} \cam{proposed} to distill knowledge from these VLMs to the detector. 
One step further, F-VLM~\cite{f-vlm} \cam{utilized} a visual encoder of VLM~\cite{clip} as its representation generator. 
Several works~\cite{detpro, promptdet} \cam{proposed} prompt engineering for VLM tailored for object detection.
\jhaaai{Although the above methods achieved state-of-the-art results, they still exhibit limited generalization to the novel classes since their detectors are supervised with the base classes only. Conversely, our proposed method additionally leverages the supervision with respect to the proxy-novel classes via class-wise mixup between a pair of base classes, improving generalization on the overall distribution of novel classes.}

% Even though they achieved state-of-the-art results, some methods~\cite{vild, detpro} require intricate preliminary processes such as obtaining all the image crops for distillation~\cite{vild} and grouped prompt learning~\cite{detpro}. 
% Conversely, our proposed method does not demand any of these computationally burdensome preparatory steps. 
% Instead, we only leverage online supervision with respect to the proxy-novel classes, which can be easily obtained through our class-wise mixup strategy.

\paragraph{Open-vocabulary object detection with pseudo-labeling} 
 Another line of works~\cite{detic, promptdet, vl-plm, pb_ovd, rkd, vldet} \cam{proposed} to pseudo-label on inexpensive datasets, which do not contain \jhaaai{bounding box annotations}, but cover lots of \jhaaai{novel} classes, to improve the performance on novel classes. Detic~\cite{detic} \cam{proposed} to use image-level labeled dataset~\cite{ridnik2021imagenet} and pseudo-label the bounding box using the max-size proposal presented from RPN~\cite{faster_rcnn} where its class label is given by image-level labels. A similar approach~\cite{rkd} further \cam{refined} the pseudo-labeling method via the class-agnostic object detector~\cite{mvit}. On the other hand, several works~\cite{regionclip, vldet} \cam{exploited} pseudo-labeling on image-caption pair dataset~\cite{cc3m} to manipulate VLM to be aware of the relationship between object and \jhaaai{novel} vocabularies within the caption. However, these pseudo-labeling methods exhibit \cam{limited generalization performance} on \cam{the truly unseen} novel classes that were not pseudo-labeled, while our method largely improves performance on these genuine novel ones, \cam{as shown in} Table \ref{table:table1_nonovl_ap}.
\begin{figure*}[t]
    \centering
    \includegraphics[width=1\linewidth]{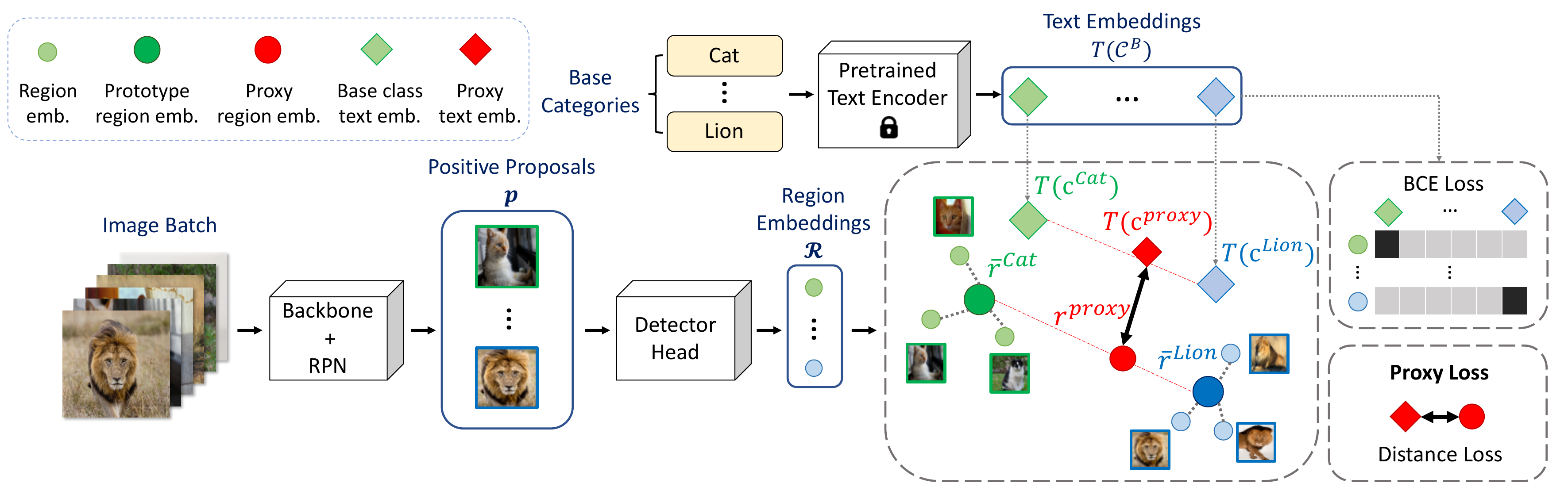}

    \caption{Overview of training pipeline for ProxyDet. Our proposed proxy loss supervises the detector to learn image-text modality with regard to proxy-novel classes via class-wise mixup between a pair of base classes.
    }

    \label{fig:overview}
    \vspace{-1mm}
\end{figure*}
%--------------------------------------------------------------------------
\section{Method}

We briefly review the overall framework of OVOD methods with notation clarification in Section \ref{method:notation_3_1}. In Section \ref{method:proxydet_3_2}, we present our method, ProxyDet, which ventures into an unexplored embedding space via proxy-novel classes, \jhaaai{enhancing the classification ability on the novel classes.}
% method 세부 섹션
% 3.1 - Overview
% Vanilla OVD (VilD Text, zeroshot clip의 활용)
%     1. Traidiontal closed-set object detector 세팅 formulation & open-set obect detector formulation 설명
%     2. mixup 알고리즘 figure 설명. 간단한 설명 및 목적. 그리고 Section 3.2, 3.3에서 어떤것을 다룰것인지? 에 대한 설명.

\subsection{Notations and Preliminaries}
\label{method:notation_3_1}
An object detector is trained on a dataset $D = \{(I_i, g_i)\}_{i=1}^{|D|}$ where $I_i$ refers to an $i$-th image and $g_i = \{(b_k, c_k)\}_{k=1}^{|g_i|}$ denotes a set of ground-truth object labels of the $i$-th image, consisting of a bounding box ($b_k$) and the corresponding class label ($c_k \in \mathcal{C}^{\text{B}} $) where $\mathcal{C}^{\text{B}}$ is the entire set of base classes. At the inference phase, OVOD aims to detect not only the base classes $\mathcal{C}^{\text{B}}$ but also novel classes $\mathcal{C}^{\text{N}}$ which satisfy $\mathcal{C}^{\text{B}} \cap \mathcal{C}^{\text{N}} = \emptyset$. 

In OVOD training, RPN~\cite{centernet2} generates a set of region proposals $\bm{P}$ for the images in the batch and positive proposals, $\bm{p}\subset\bm{P}$, with sufficient IoUs with ground-truth bounding boxes of base classes are selected and \cam{labeled according to the class of the nearest ground-truth box}. Then, each positive proposal $p \in \bm{p}$ is vectorized into a region embedding $r \in \mathbb{R}^{M}$ through detector head with RoI-Align~\cite{mask_rcnn}, forming a set of region embeddings $\bm{\mathcal{R}}$. For classification training of base class objects, \jhaaai{we adopt} binary cross entropy (BCE) loss~\cite{detic} using cosine similarity between $r$ and text embeddings of base classes $\mathcal{T}(\mathcal{C}^{\text{B}}) \in \mathbb{R}^{|\mathcal{C}^{\text{B}}| \times M}$, where $\mathcal{T}(\cdot)$ denotes text encoder of \jhaaai{CLIP}~\cite{clip}.

%we obtain a set of proposals $\bm{P}$ for all the images in the batch via regional proposal network (RPN)~\cite{centernet2}. Among the proposals, we obtain a set of positive proposals $\bm{p}$ ($\bm{p} \subset \bm{P}$) whose IoUs with regard to the nearest ground-truth bounding box $b^{*}$ are higher than a threshold (e.g., 0.5). Each of positive proposal $p \in \bm{p}$ is then vectorized into regional embedding $r \in \bm{\mathcal{R}}$ through detector head with ROI-Align~\cite{mask_rcnn}, where $\mathcal{R}$ is the set of all the regional embeddings in the batch and $r \in \mathbb{R}^{M}$. For classification training of base class objects, we applied binary cross entropy (BCE) loss~\cite{detic} using cosine similarity between $r$ and text embedding of base classes $\mathcal{T}(\mathcal{C}^{\text{B}}) \in \mathbb{R}^{|\mathcal{C}^{\text{B}}| \times M}$.
% \begin{equation}

% \end{equation}

\subsection{ProxyDet: Learning Proxy Novel Classes via Class-wise Mixup}
\label{method:proxydet_3_2}
% To facilitate the model's ability to recognize new categories during test phase, it is crucial to expand the boundary of the visual and textual representations towards novel classes. 
% To achieve this, we synthesize the proxy-novel classes by separately mixing the visual and textual class \emph{prototype} of the base classes, and then regulate them to be close to each other.
In order to facilitate the detector to recognize novel classes not seen during the training phase, it is crucial for the detector to explore the \hskim{embedding} space \jhaaai{that is proximate to the space of novel classes}. \jhaaai{Inspired by our observation that a number of novel classes reside within the convex hull delineated by the base classes (Figure \ref{fig:tsne}), we synthesize the proxy-novel classes via class-wise mixup between the representations of the base classes.}
\jhaaai{After obtaining} the visual and textual \hskim{embedding} of these proxy-novel classes, \jhaaai{we} regulate them to be close to each other. The overall framework of our method is illustrated in Figure~\ref{fig:overview}.
% the visual (and textual) representations between base classes, some of which are close to novel classes as seen in~\figureautorefname~\ref{fig:novel_similarity}. Then, we regulate the visual and textual embeddings of these proxy-novel classes to be close to each other.}
% \sout{\hskim{To achieve this, we introduce the synthesis of proxy-novel classes, which consist of visual and textual embeddings that are generated from a model and a VLM, respectively, some of which are close to novel classes as seen in~\figureautorefname~\ref{fig:novel_similarity}}. 
% \hskim{Then, we regulate the visual and textual embeddings of these proxy-novel classes to be close to each other.}} %as same as we actually do on base classes at detector's classification head~\cite{faster_rcnn}.

%To achieve this, we synthesize the proxy-novel classes by obtaining the mixed representations for each of visual and textual class prototypes belonging to the seen classes, and then regulate them to be close to each other.
%Visual and Textual Prototype.
% \vspace{-4mm}
\paragraph{Synthesizing proxy-novel classes}
Given the visual and
textual \jhaaai{prototypes} of base classes in a batch, $\bar{r}^{\text{B}}$ and $\mathcal{T}(\mathcal{C}^{\text{B}})$, we randomly sample a pair of prototypes $\big(\bar{r}^{\text{B}_{i}}$,  $\mathcal{T}(c^{\text{B}_{i}})\big)$ and $\big(\bar{r}^{\text{B}_{j}}$,  $\mathcal{T}(c^{\text{B}_{j}})\big)$ that contain class-specific features of base \cam{classes} $c^{\text{B}_{i}}$ and $c^{\text{B}_{j}}$, respectively. The procedure for constructing the prototype will be elaborated on \jhaaai{afterwards}.

% Given the visual and textual class prototypes, \hskim{which contain class-specific features of base class $c^{\text{B}_{i}}$ and $c^{\text{B}_{j}}$, respectively,} of base classes in a batch, $\bar{r}^{\text{B}}$ and $\mathcal{T}(\mathcal{C}^{\text{B}})$.  
% \hskim{The procedure for constructing the prototype will be elaborated subsequently.}
% \hskim{\sout{We randomly sample a pair of prototypes $\big(\bar{r}^{\text{B}_{i}}$,  $\mathcal{T}(c^{\text{B}_{i}})\big)$ and $\big(\bar{r}^{\text{B}_{j}}$,  $\mathcal{T}(c^{\text{B}_{j}})\big)$.}} \hskim{\sout{that contain class-specific features of base class $c^{\text{B}_{i}}$ and $c^{\text{B}_{j}}$, respectively. The procedure for constructing the prototype will be elaborated subsequently.}}

% In order to obtain the visual and textual representations of proxy-novel class $c^{\text{proxy}} \in \mathcal{C}^{\text{proxy}}$, we propose a class-wise mixup approach between these prototypes of base classes that contain class-specific features:
In order to obtain the visual and textual representations of proxy-novel class 
% $c^{\text{proxy}} \in \mathcal{C}^{\text{proxy}}$, C_proxy set이라는게 애매한데, C를 base와 novel 그 이상의 모든 클래스들의 집합이라고 표현하면 어떨까요? 개념적으로는 맞는 것 같은데 톤이 너무 세질까요?
\jyyoo{$c^{\text{proxy}} \in \mathcal{C}$, where $\mathcal{C}$ is the set of all classes beyond the base and novel classes,}
we propose a class-wise mixup approach between these prototypes of base classes:
% \hskim{a pair of randomly sampled prototypes of base classes, $\big(\bar{r}^{\text{B}_{i}}$,  $\mathcal{T}(c^{\text{B}_{i}})\big)$ and $\big(\bar{r}^{\text{B}_{j}}$,  $\mathcal{T}(c^{\text{B}_{j}})\big)$}: 
% \hskim{\sout{This approach is supported by the observation from the \figureautorefname~\ref{fig:novel_similarity}}}.
% The formulation for our class-wise mixing strategy is presented as follows:
% This proposal is supported by the observation from Figure \ref{fig:novel_similarity}.
% Explanation on how to construct a prototype will be provided subsequently.
%To obtain the visual and textual representations of proxy-novel class $c^{\text{proxy}} \in \mathcal{C}^{\text{proxy}} $, we present class-wise mixup between \emph{prototype} embeddings of base classes that contain class-specific features, based on the observations from Figure \ref{fig:novel_similarity}:
\begin{equation}
\begin{split}
    r^{\text{proxy}} &= \mathcal{V}\big(\lambda \bar{r}^{\text{B}_{i}} + (1 - \lambda) \bar{r}^{\text{B}_{j}}\big) \\
     \mathcal{T}(c^{\text{proxy}}) &= \mathcal{V}\big(\lambda \mathcal{T}(c^{\text{B}_{i}}) + (1 - \lambda) \mathcal{T}(c^{\text{B}_{j}})\big)
\end{split}
\label{eq:mixup}
\end{equation}
where $r^{\text{proxy}}$ and $\mathcal{T}(c^{\text{proxy}})$ refer to the region (i.e., visual) and textual prototype embedding of $c^{\text{proxy}}$, respectively. \cam{$\mathcal{V}(\mathbf{x})=\frac{\mathbf{x}}{||\mathbf{x}||_{2}}$ refers to the L2 normalization of a vector $\mathbf{x}$} and $\lambda $ follows $\text{Beta}(\gamma, \gamma)$.

%We linearly interpolate a randomly-sampled pair of base class prototypes $\big(\bar{r}^{\text{B}_{i}}$,  $\mathcal{T}(c^{\text{B}_{i}})\big)$ and $\big(\bar{r}^{\text{B}_{j}}$,  $\mathcal{T}(c^{\text{B}_{j}})\big)$.
%$\mathcal{T}(c^{\text{B}_{i}})$ and $\bar{r}^{\text{B}_{i}}$ denote the prototypical text and region embedding of base class $c^{\text{B}_{i}} \in \mathcal{C}^{\text{B}}$, respectively. 

%and $c^{\text{B}_{i}}, c^{\text{B}_{j}}$ are random-uniformly sampled from one of base classes appeared within the image batch. 
% \vspace{-3mm}
\paragraph{Constructing prototype} The way of constructing visual and textual prototypes is a crucial factor in obtaining sophisticated representations of proxy-novel classes. Textual prototypes can be obtained by prompting the category name~$c^{\text{B}_{i}}$ into the text encoder of a pretrained VLM~\cite{clip}. On the other hand, obtaining class-specific region embedding~$\bar{r}^{\text{B}_{i}}$ is more complicated as there are multiple region embeddings \cam{from a batch of images} that have the potential to represent a class~$c^{\text{B}_{i}}$. A naive solution is to calculate the centroid of all these region embeddings which correspond to $c^{\text{B}_{i}}$:
%Although class-specific text embeddings $\mathcal{T}(c^{\text{B}_{i}})$ can be easily obtained by forwarding the category name of $c^{\text{B}_{i}}$ through largely pretrained text encoder $\mathcal{T}$, acquiring class-specific region embedding $\bar{r}^{\text{B}_{i}}$ is non-trivial since there are a number of region embeddings that can represent class $c^{\text{B}_{i}}$. 
%A simple and naive solution is to estimate the centroid of all the positive region embeddings where the class labels of their closest ground-truth box are all equal to $c^{\text{B}_{i}}$:
\begin{equation}
\begin{split}
    & \bar{r}^{\text{B}_{i}} = \mathcal{V}\big(\sum_{r \in \bm{\mathcal{R}^{\text{B}_{i}}}} F(r) * r\big), \\
    & F(r) = \frac{1}{|\bm{\mathcal{R}^{\text{B}_{i}}}|}
\end{split}
\label{eq:prototype}
\end{equation}
where $\bm{\mathcal{R}^{\text{B}_{i}}}$ refers to the set of region embeddings for positive proposals \cam{labeled as} $c^{\text{B}_{i}}$. $F(r)$ denotes the weighting function that determines the contribution of each positive region embedding $r$ towards constructing the prototype. In Eq.~\ref{eq:prototype}, all the region embeddings are given equal weight in the construction process.

\begin{figure}[t]
    \centering
    \begin{subfigure}[b]{0.23\textwidth}
        \centering
        \includegraphics[width=\textwidth]{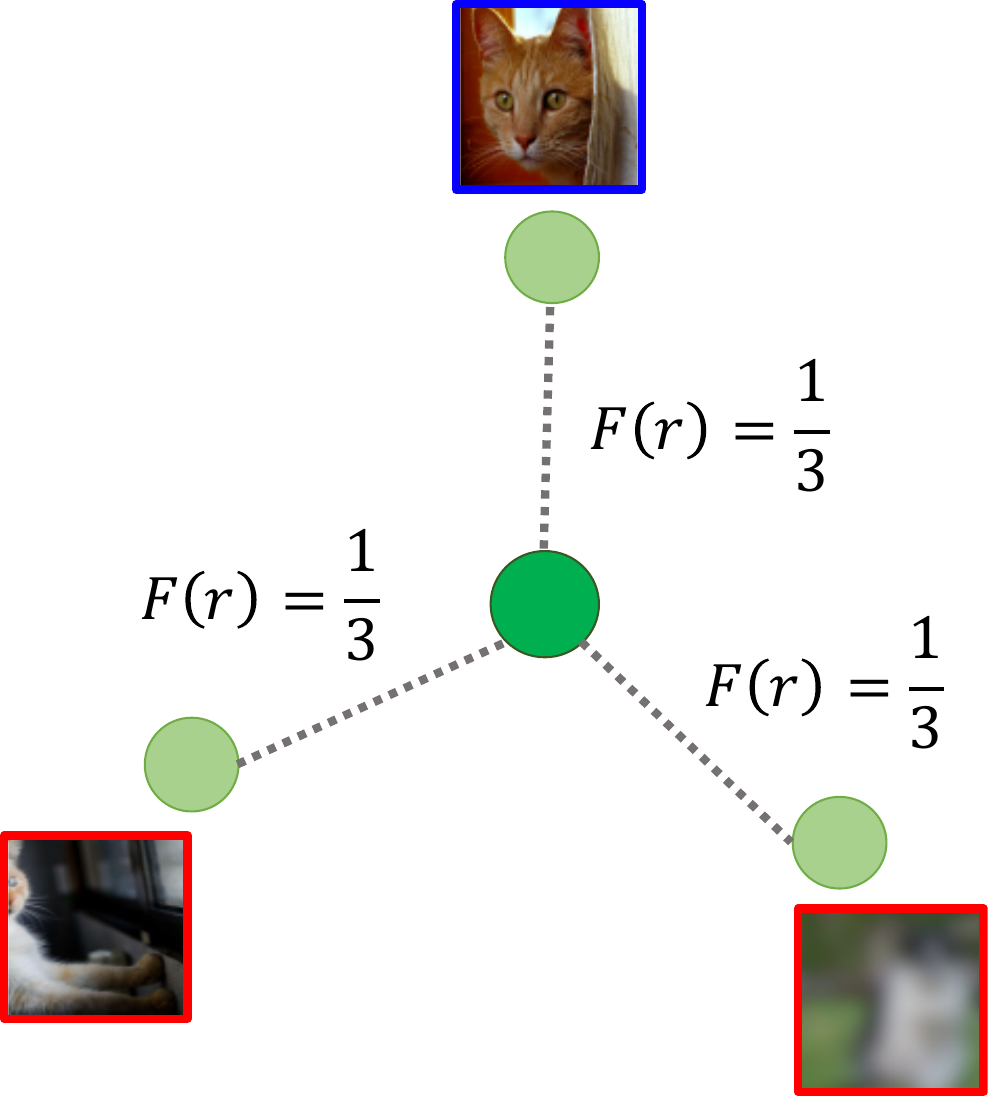}
        \caption{Centroid prototype}
        \label{fig:averaged_prototype}
    \end{subfigure}
    % \hspace{10mm}
    \hfill
    \begin{subfigure}[b]{0.23\textwidth}
        \centering
        \includegraphics[width=\textwidth]{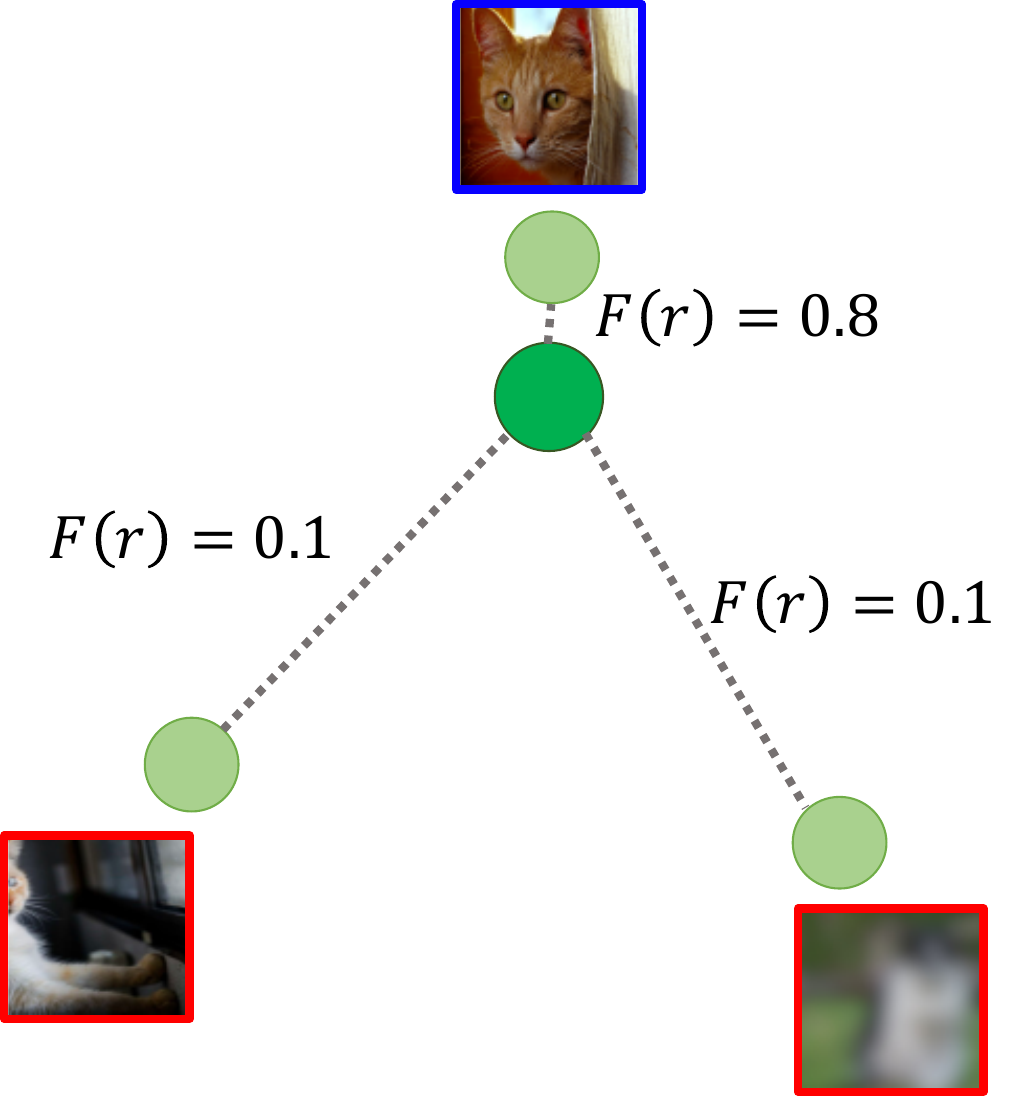}
        \caption{Quality-aware prototype}
        \label{fig:weighted_prototype}
    \end{subfigure}
    \caption{Construction of \jhaaai{visual} prototype embedding. Red boxes denote low-quality proposals (e.g., partially localized, blurry) while blue boxes denote high-quality proposals. Quality-aware prototype reduces the influence of low-quality proposals, resulting in robust prototype.
    }
    \label{fig:prototype}
    \vspace{-3mm}
\end{figure}

% \vspace{-3mm}
\paragraph{Robust \jhaaai{visual} prototype} However, simply averaging all region embeddings to construct a centroid prototype embedding could be sub-optimal since the positive proposals have the potential to include low-quality localizing results as shown in Figure~\ref{fig:prototype}. To address this issue, we propose an approach to construct a robust prototype by incorporating the weighting function $F(r)$ with a measure that is aware of the quality of proposals, denoted as $\phi(\cdot)$.
\begin{equation}
\begin{split}
    F(r) = \frac{\text{exp}\big(\phi(r)\big)}{\sum_{\hat{r} \in \bm{\mathcal{R}^{\text{B}_{i}}}} \text{exp}\big(\phi(\hat{r})\big)}
\end{split}
\label{eq:prototype_weighting_function}
\end{equation}
For designing the proposal quality-aware measure function $\phi(\cdot)$, we leverage \emph{objectness score} from RPN or \emph{IoU} with the nearest ground-truth box. By doing so, our proxy-novel synthesis in Eq.~\ref{eq:mixup} is conditioned on more robust prototype embeddings which can more precisely estimate the class-wise representations with respect to the base classes.

%Through Eq~\ref{eq:prototype_weighting_function}, our proxy-novel synthesis (Eq (\ref{eq:mixup})) is conditioned on more robust prototype embeddings which can more accurately estimate the class-wise representations with respect to the base classes.

%--------------------------------------------------------------------------
%--------------------------------------------------------------------------
% \vspace{-4mm}
 \paragraph{Proxy loss} 
% 위 목적을 달성하기 위하여, $r^{proxy}~= r^{novel}$이 proxy-novel text embedding $T~\mathcal{C}^{\text{N}}$ (which is approximately novel as show in figure 2)에 가까워지도록 학습할 수 있도록 한다. 
Building upon the visual and textual representations of synthesized proxy-novel classes with the robust formulation of prototype embedding, we derive our proposed training objective as follows:
\begin{equation}
    \mathcal{L}_{\text{proxy}} = || \mathcal{T}(c^{\text{proxy}}) - r^{\text{proxy}} ||_{1}
\label{eq:proxy_loss}
\end{equation}
Intuitively, we train the OVOD model by regularizing the region embeddings of proxy-novel classes to minimize its distance with the proxy-novel text embeddings, which effectively encourages the model to explore the proximal \hskim{embedding} space of the novel classes.
 \paragraph{Inference} As our proxy loss (Eq.~\ref{eq:proxy_loss}) is designed to improve \cam{the} classification performance of novel classes, the objective is different from the conventional BCE loss which prioritizes \cam{the classification} performance of base classes.
 \jhaaai{\cam{To} prevent the potential contention between these two losses, we apply each loss separately on two separate detector heads with identical architecture.}
 %Since our proxy loss Eq (\ref{eq:proxy_loss}) aims to enhance classification performance towards novel classes while conventional BCE loss focuses on that of base classes, there could be contention.
 % Therefore, we apply each loss separately on two separate detector heads with identical architecture. 
 \cam{During the inference phase, we compute classification scores of the two region embeddings from the separated heads by calculating cosine similarity against the text embeddings of all the classes. Then, we merge the two classification scores by geometric mean, following~\citet{vild} and \citet{f-vlm}:}
% produced by RPN and detection head, and classify each one by computing dot product similarity against the text embeddings of all the classes and selecting the class with the highest similarity. For calculating the classification score of a region embedding,} we merge the classification scores obtained from the two heads by geometric mean, following~\citet{vild} and \citet{f-vlm}:
\begin{equation}
    s_i = 
\begin{cases}
    \big(r w^{\top}\big)^{\alpha}_{i} * \big(r' w^{\top}\big)^{1-\alpha}_{i}, & \text{if } c^i \in \mathcal{C}^{\text{B}} \\
    \big(r w^{\top}\big)^{\beta}_{i} * \big(r' w^{\top}\big)^{1-\beta}_{i}, & \text{if } c^i \in \mathcal{C}^{\text{N}}
\end{cases}
\label{eq:score_fusion}
\end{equation}
where $\ s_i$ is the final classification score for $i$-th class $c^i$, classifier weight $w$ is \cam{L2 normalized text embeddings from} $\mathcal{T}(\mathcal{C}^{\text{B}} \cup \mathcal{C}^{\text{N}})$, and \cam{$r$,$r'$} are \cam{L2 normalized} region \cam{embeddings} from \cam{the} head trained with proxy loss and the other head with BCE loss, respectively.  
$\alpha, \beta \in [0, 1]$ determines the ratio of reflecting classification score from proxy loss head for base/novel classes, respectively.

\section{Experiments}
\label{exp}

\subsection{Datasets}
\label{exp:datasets} 
We evaluate our method on the widely-used OVOD benchmark datasets, LVIS~\cite{gupta2019lvis}, COCO~\cite{coco}, and Objects365~\cite{object365}, to compare the performance against the previous works. 
% \vspace{-4mm}

\noindent{\textbf{LVIS and ImageNet-LVIS.}}
% The LVIS dataset is composed of 100k images with annotation encompassing bounding boxes and segmentation mask labels with respect to 1203 classes. 
\jyyoo{The LVIS dataset consists of 100k annotated images with bounding boxes and segmentation mask labels for 1203 classes.}
The object classes in LVIS are divided into three distinct groups, namely \textit{"frequent"}, \textit{"common"}, and \textit{"rare"}, based on the occurrence frequency within the dataset. We follow Open-Vocabulary LVIS (OV-LVIS) benchmark~\cite{vild} that uses common and frequent class groups for base classes (866 categories) and the rare group for novel classes (337 categories). \cam{Optionally,} we utilize ImageNet-LVIS (IN-L) as an additional \cam{training} dataset with image-level labels \cite{detic}, which is a subset of ImageNet-21k~\cite{ridnik2021imagenet} containing only the categories that intersect with those of LVIS.
% For the case where using ImageNet-LVIS (IN-L)~\cite{detic} as additional training source, we only apply our proposed proxy loss on LVIS dataset.

 \noindent{\textbf{COCO and COCO Caption.}}
We adopt Open-Vocabulary COCO (OV-COCO) benchmark proposed in~\citet{zeroshot-obj} and \citet{ovrcnn}, which divides COCO classes into base classes (48 categories) and novel classes (17 categories). We additionally use \jhaaai{captions} in COCO dataset as image-level labels, following~\citet{detic}. 
%We only applied our proxy loss using box-level labels.

% \paragraph{Object365 and OpenImages.}
% We utilize Object365~\cite{object365} and OpenImages~\cite{openimages} for cross-datasets evaluation.
% Object365 has 365 categories over 2M images and OpenImages contains 500 categories over 1.8M images. 
% \ctv{Although they are willing to show the effectiveness of our models across the datasets, some of their categories overlap with base categories.
% Therefore, we do our best to show both of the overlapped and non-overlapped performance.}
% 위에가 안되면, 아래 문단으로 대체 예정
\noindent{\textbf{Objects365.}}
Objects365 has 365 categories over 2M images. We use Objects365~\cite{object365} for cross-dataset evaluation to substantiate the generalization capability.
% For both Objects365 and OpenImages datasets, we further report AP for rare categories whose appearance frequency falls within the lower third.
% To substantiate the generalization capability, we conduct experiments of the cross-dataset evaluation  where a model trained with LVIS is evaluated on the other widely-used detection datasets, Object365 and OpenImages.
% Object365 has 365 categories over 2M images and OpenImages contains 500 categories over 1.8M images. For Object365 and OpenImages, we additionally report AP for novel categories defined within each dataset.

% We conduct the experiments to evaluate the open-vocabulary property where a model trained with LVIS is evaluated on other detection datasetscontaining difference category sets (COCO~\cite{coco}, Object365~\cite{object365}, OpenImages~\cite{openimages}). 
% For Object365 and OpenImages, we additionally report AP for novel categories defined within each dataset.

\subsection{Implementation Details}
\label{exp:impl_details}

In OV-LVIS setup, we follow the detector model architecture of Detic~\cite{detic} and its training recipe.
Our object detection model is based on Cascade Mask-RCNN~\cite{cascade_rcnn} with ResNet50~\cite{resnet} or Swin-B~\cite{liu2022swin} backbone, employing CenterNet2~\cite{centernet2} as RPN. For obtaining the text embedding for the classifier, we use CLIP ViT-B/32~\cite{clip}. 
The training recipe using ResNet50 backbone consists of federated loss~\cite{centernet2} and binary cross-entropy loss. 
We \cam{trained} the detector for 90,000 iterations (1x schedule) with 64 batch size and AdamW~\cite{loshchilov2017decoupled} optimizer. For the learning rate schedule, we increased 0 to 2e-4 for the first 1000 warmup iterations and decayed by \cam{cosine annealing strategy~\cite{loshchilov2016sgdr}.} %a factor of 10 at 30,000 iterations.
In the case of using image-level supervision, we fine-tune the pretrained detector whose weight parameters \cam{were} trained on LVIS base classes, and thus total training iteration is 2x schedule.
% For each of mini-batch, the ratio of detection dataset (i.e., LVIS) and image-level labeled dataset (i.e., IN-L) is 1:4.
For the image-level labeled dataset, IN-L, we do not supervise the detector with our proposed proxy loss.
% Note that image-level labeled dataset is not used for generating proxy-novel classes since we figure out that pseudo-\jyyoo{labeled}
% annotated 
% bounding boxes~\cite{detic} are not precisely localized enough to be \hskim{\sout{class representative} a prototype} for our proposed class-wise mixup strategy.
\jhaaai{For evaluation, we report mask AP for all the classes and $\text{AP}_r$ for the \cam{rare (novel)} classes, respectively.
$\text{AP}_r$ is the \cam{main metric} we \cam{primarily concentrate on.}}
% $\text{AP}_r$ is the \cam{main metric} we \cam{evaluate with.}}
% \jhaaai{For evaluation, we report mask AP for novel classees, $\text{AP}_r$, as the main benchmark metric.}

In OV-COCO setup, we adopt Faster R-CNN~\cite{faster_rcnn} with ResNet50-C4 backbone. We \cam{trained} for 1x schedule with batch size 16 and SGD optimizer.
\jhaaai{For evaluation, we report box $\text{AP}_{50}$ for the base ($\text{AP}_{\text{base}}$), novel ($\text{AP}_{\text{novel}}$), and all classes ($\text{AP}$), where $\text{AP}_{\text{novel}}$ is the main metric.}
% For evaluation, we \jhaaai{mainly report box $\text{AP}_{\text{novel}}$ at IoU threshold 0.5.}
% \cam{For evaluation of Objects365, we additionally evaluate AP $\text{AP}^{\text{rare}}_{50}$ }
For the hyper-parameters of our method, we random-uniformly sample the mixing coefficient $\lambda$ (in Eq.~\ref{eq:mixup}), following $\text{Beta}(1, 1)$ distribution. For the score fusion in Eq.~\ref{eq:score_fusion}, we use $\alpha=0.45$ and $\beta=0.65$. % We report the results as an average over 3 runs.

% mixup-specific parameters
% LVIS에서의 setup
% coco에서의 setup

\begin{table}[t]

\centering
\smallskip\noindent
\tabcolsep=0.07cm
% \vspace{-1mm}
% ============ APr, AP only =============
% \caption{Performance comparison on Open-Vocabulary LVIS benchmark. \jhaaai{LVIS-base denotes training with base classes only, while $^*$ denotes fully supervised with the base and novel classes.}
% % $\dag$ refers to additional using pseudo-box labels w.r.t. novel classes.
% For fair comparison, all the methods employed ResNet50 backbone.}
% \resizebox{1\linewidth}{!}{
% \begin{tabular}{lccc}
% \toprule
% Method & Training Source & $\text{AP}_r$ & \textcolor{gray}{AP} \\ 
% \midrule
% ViLD~\cite{vild} & LVIS-base & 16.6 & \textcolor{gray}{25.5} \\
% RegionCLIP~\cite{regionclip} & LVIS-base & 17.1 & \textcolor{gray}{28.2} \\
% F-VLM~\cite{f-vlm} & LVIS-base & 18.6 & \textcolor{gray}{24.2} \\
% % Detpro~\cite{detpro} & LVIS & 19.8 & 25.9 \\
% ProxyDet (ours) & LVIS-base & \textbf{18.9} & \textcolor{gray}{30.1} \\
% \midrule
% \textcolor{gray}{Fully-supervised$^{*}$} & \textcolor{gray}{LVIS} & \textcolor{gray}{25.5} & \textcolor{gray}{31.1}  \\
% % VLDet~\cite{vldet} & LVIS + CC3M & 21.7 & 30.1 \\
% % Detic~\cite{detic} & LVIS + CC3M & 19.5 & 30.9 \\
% Detic~\cite{detic} & LVIS-base + IN-L & 24.6 & \textcolor{gray}{32.4} \\
% % RKD$\dag$~\cite{rkd} & LVIS + IN-L & 25.2 & \textcolor{gray}{32.9} \\
% ProxyDet (ours) & LVIS-base + IN-L & \textbf{26.2} & \textcolor{gray}{32.5} \\

% ============ APr, AP only with backbone =============
\resizebox{1\linewidth}{!}{
\begin{tabular}{lcccc}
\toprule
Method & Backbone & Training source & $\text{AP}_r$ & \textcolor{gray}{AP} \\ 
\midrule
ViLD~(\citeauthor{vild}) & R50 & LVIS-base &  16.6 & \textcolor{gray}{25.5} \\
RegionCLIP~(\citeauthor{regionclip}) & R50 & LVIS-base & 17.1 & \textcolor{gray}{28.2} \\
F-VLM~(\citeauthor{f-vlm}) & R50 & LVIS-base & 18.6 & \textcolor{gray}{24.2} \\
% Detpro~\cite{detpro} & LVIS & 19.8 & 25.9 \\
\textbf{ProxyDet (ours)} & R50 & LVIS-base & \textbf{18.9} & \textcolor{gray}{30.1} \\
\midrule
\textcolor{gray}{Fully-supervised$^{*}$} & \textcolor{gray}{R50} & \textcolor{gray}{LVIS} & \textcolor{gray}{25.5} & \textcolor{gray}{31.1}  \\
% VLDet~\cite{vldet} & LVIS + CC3M & 21.7 & 30.1 \\
% Detic~\cite{detic} & LVIS + CC3M & 19.5 & 30.9 \\
Detic~(\citeauthor{detic}) & R50 & LVIS-base + IN-L & 24.6 & \textcolor{gray}{32.4} \\
% RKD$\dag$~\cite{rkd} & LVIS + IN-L & 25.2 & \textcolor{gray}{32.9} \\
\textbf{ProxyDet (ours)} & R50 & LVIS-base + IN-L & \textbf{26.2} & \textcolor{gray}{32.5} \\
% RKD$\dag$~\cite{rkd} & LVIS + IN-L & 25.2 & \textcolor{gray}{32.9} \\
Detic~(\citeauthor{detic}) & Swin-B & LVIS-base + IN-L & 33.8 & \textcolor{gray}{40.7} \\
\textbf{ProxyDet (ours)} & Swin-B & LVIS-base + IN-L & \textbf{36.7} & \textcolor{gray}{41.5} \\

% ============ APr, APc, APf, AP  =============
% \begin{tabular}{lccccc}
% \toprule
% Method & Training Source & $\text{AP}_r$ & \textcolor{gray}{$\text{AP}_c$} & \textcolor{gray}{$\text{AP}_f$}& \textcolor{gray}{\text{AP}} \\ 
% \midrule
% ViLD~\cite{vild} & LVIS-base & 16.6 & \textcolor{gray}{24.6} & \textcolor{gray}{30.3} & \textcolor{gray}{25.5} \\
% RegionCLIP~\cite{regionclip} & LVIS-base & 17.1 & \textcolor{gray}{27.4} & \textcolor{gray}{34.0} & \textcolor{gray}{28.2}   \\
% F-VLM~\cite{f-vlm} & LVIS-base & 18.6 & \textcolor{gray}{-} & \textcolor{gray}{-} & \textcolor{gray}{24.2} \\
% % Detpro~\cite{detpro} & LVIS & 19.8 & 25.9 \\
% \textbf{ProxyDet (ours)} & LVIS-base & \textbf{18.9} & \textcolor{gray}{30.5} & \textcolor{gray}{34.6} & \textcolor{gray}{30.1} \\
% \midrule
% \textcolor{gray}{Fully-Supervised$^{*}$} & \textcolor{gray}{LVIS} & \textcolor{gray}{25.5} & \textcolor{gray}{30.4} & \textcolor{gray}{35.2} & \textcolor{gray}{31.1}  \\
% % VLDet~\cite{vldet} & LVIS + CC3M & 21.7 & 30.1 \\
% % Detic~\cite{detic} & LVIS + CC3M & 19.5 & 30.9 \\
% Detic~\cite{detic} & LVIS-base + IN-L & 24.6 & \textcolor{gray}{32.5} & \textcolor{gray}{35.6} & \textcolor{gray}{32.4} \\
% % \textcolor{gray}{Object-Centric$\dag$~\cite{rkd}} & \textcolor{gray}{LVIS-base + IN-L} & \textcolor{gray}{25.2} & \textcolor{gray}{33.4} & \textcolor{gray}{35.8} & \textcolor{gray}{32.9} \\
% \textbf{ProxyDet (ours)} & LVIS-base + IN-L & \textbf{26.2} & \textcolor{gray}{32.4} & \textcolor{gray}{35.4} & \textcolor{gray}{32.5} \\

\bottomrule

\end{tabular}
}
\vspace{-2mm}
\caption{Performance comparison on Open-Vocabulary LVIS benchmark. \jhaaai{LVIS-base denotes training with base classes only, while $^*$ denotes fully supervised with the base and novel classes. R50 denotes ResNet50.}
% $\dag$ refers to additional using pseudo-box labels w.r.t. novel classes.
}
\vspace{-5.5mm}
\label{table:table2_ov_lvis}
% \vspace{5.5mm}
\end{table}

\begin{table}[t]
\begin{center}
\small
\setlength\tabcolsep{5pt}
% \vspace{0.5mm}
% \vspace{-0.25mm}
\vspace{1mm}
\resizebox{1\linewidth}{!}{
\begin{tabular}{lccccc}
\toprule
Method & $\text{AP}_{\text{novel}}$ & \textcolor{gray}{$\text{AP}_{\text{base}}$} & \textcolor{gray}{AP} \\
\midrule
WSDDN~\cite{wsddn}                & 19.7 & \textcolor{gray}{19.6} & \textcolor{gray}{19.6} \\
Cap2Det~\cite{cap2det}            & 20.3 & \textcolor{gray}{20.1} & \textcolor{gray}{20.1} \\
OVR-CNN~\cite{ovrcnn} & 22.8 & \textcolor{gray}{46.0} & \textcolor{gray}{39.9} \\
ViLD$\dag$~\cite{vild} & \textcolor{black}{27.6} & \textcolor{gray}{59.5} & \textcolor{gray}{51.3} \\
RegionCLIP$\ddag$~\cite{regionclip} & 26.8 & \textcolor{gray}{54.8} & \textcolor{gray}{47.5} \\                                    
Detic~\cite{detic} & 27.8 & \textcolor{gray}{47.1} & \textcolor{gray}{45.0} \\
\textbf{ProxyDet (ours)} & \textbf{30.4} & \textcolor{gray}{52.6} & \textcolor{gray}{46.8} \\
\bottomrule                      
\end{tabular}
}
\vspace{-2.5mm}
\end{center}
\caption{Performance comparison on Open-Vocabulary COCO benchmark. $\dag$ denotes using 8x training schedule. \\ \cam{$\ddag$ denotes leveraging an additional pretraining scheme for the visual encoder with extensive object concepts.}}
\label{table:table3_ov_coco}
\vspace{-3mm}
\end{table}
\subsection{Main Results}
\label{exp:main_results}
% \vspace{-2mm}
\noindent{\textbf{Open-vocabulary LVIS.}} 
In Table~\ref{table:table2_ov_lvis}, we compare our proposed method, ProxyDet, with the other state-of-the-art OVOD methods on OV-LVIS benchmark. 
When only LVIS-base is used for training without any external dataset, our method shows the best performance on novel classes, $18.9\ \text{AP}_r$. This result supports the fact that ProxyDet flourishes open-vocabulary property without any other additional training source.
However, the fully-supervised detector trained with all the categories including base and novel classes (5th row) provides that there is still large headroom for improvement, achieving $25.5\ \text{AP}_r$. 
Notably, with image-level supervision from IN-L, our model (7th row) outperforms the fully-supervised detector, reaching $26.2\ \text{AP}_r$. 
Moreover, compared to Detic which also utilizes IN-L, our method outperforms them by a large margin, achieving +$1.6\ \text{AP}_r$ and +$2.9\ \text{AP}_r$ improvement when using ResNet50 and Swin-B backbones, respectively.
% We note that although our method does not leverage refined pseudo box annotations from the massively-pretrained RPN~\cite{mvit} as in Object-Centric~\cite{rkd}, our approach outperforms it in terms of novel class performance, indicating the effectiveness of our proxy loss technique.

\noindent{\textbf{Open-vocabulary COCO.}} 
Table~\ref{table:table3_ov_coco} displays the performance comparison results on OV-COCO benchmark. 
The pioneering work, OVR-CNN, shows moderate performance on novel classes via pretraining with image-caption pairs. 
With the advent of massively pretrained VLMs (i.e., CLIP), several works (ViLD, RegionCLIP, Detic) greatly enhanced novel class performance.
Notably, our proposed method further improves $\text{AP}_{\text{novel}}$ by $+2.6$ \hskim{compared to Detic}, achieving the best performance among the comparisons. This result validates the effectiveness of our proposed supervision with regard to proxy-novel classes.

\begin{table}[t]
  \small
  \centering
    \smallskip\noindent
    \tabcolsep=0.075cm
  \resizebox{1\linewidth}{!}{
  
    % with OpenImages
        % \begin{tabular}{lc|c|cc|cc}
        % \toprule
        % with OpenImages, and LVIS-base + IN-L
        % \multirow{2}{*}{Method} & \multirow{2}{*}{Training Source}&\multicolumn{1}{c|}{COCO}&\multicolumn{2}{c|}{Objects365}&\multicolumn{2}{c}{OpenImages}\\
        % & & $\text{AP}_{50}$ & $\text{AP}_{50}$ & $\text{AP}^{\text{rare}}_{50}$ & $\text{AP}_{50}$ & $\text{AP}^{\text{rare}}_{50}$ \\
        % \midrule
        % \textcolor{gray}{Target-Supervised$^{*}$} & \textcolor{gray}{Target dataset} & \textcolor{gray}{67.6} & \textcolor{gray}{38.6} & \textcolor{gray}{-} & \textcolor{gray}{64.4} & \textcolor{gray}{-} \\
        % \midrule
        % LVIS-Supervised$\S$ & LVIS-base  & 55.4 & 18.9 & 16.2 & 37.1 & 53.9 \\
        % % \midrule
        % ViLD & LVIS-base & 55.6 & 18.2 & 15.4 & 27.6 & 41.8 \\
        % DetPro & LVIS-base & 53.8 & 18.8 & 13.9 & \textbf{40.0} & 56.7 \\
        % F-VLM$\dag$ & LVIS-base & 53.1 & \textbf{19.2} & 15.4 & 37.7 & 52.8 \\
        % \textbf{ProxyDet (ours)} & LVIS-base & \textbf{57.0} & 19.1 & \textbf{16.5} & 39.4 & \textbf{58.4} \\
        % \midrule
        % Detic & LVIS-base + IN-L & 56.3 & 21.7 & 20.5 & 42.2 & 59.7 \\
    % \textbf{ProxyDet (ours)} & LVIS-base + IN-L & \textbf{56.8} & \textbf{22.2} & \textbf{20.5} & \textbf{42.5} & \textbf{61.0} \\

        \begin{tabular}{lc|c|cc}
        \toprule
        \multirow{2}{*}{Method} & \multirow{2}{*}{Training Source}&\multicolumn{1}{c|}{COCO}&\multicolumn{2}{c}{Objects365}\\
        & & $\text{AP}_{50}$ & $\text{AP}_{50}$ & $\text{AP}^{\text{rare}}_{50}$ \\
        \midrule
        \textcolor{gray}{Target-supervised$^{*}$} & \textcolor{gray}{Target dataset} & \textcolor{gray}{67.6} & \textcolor{gray}{38.6} & \textcolor{gray}{-} \\
        \midrule
        LVIS-supervised$\S$ & LVIS-base  & 55.4 & 18.9 & 16.2 \\
        % \midrule
        ViLD~\cite{vild} & LVIS-base & 55.6 & 18.2 &  \ \ 15.4$^{\dag}$ \\
        DetPro~\cite{detpro} & LVIS-base & 53.8 & 18.8 & \ \ 13.9$^{\dag}$ \\
        F-VLM~\cite{f-vlm} & LVIS-base & 53.1 & \textbf{19.2} & \ \ 15.4$^{\dag}$ \\
        \textbf{ProxyDet (ours)} & LVIS-base & \textbf{57.0} & 19.1 & \textbf{16.5} \\

        \bottomrule
    \end{tabular}
}
\caption{Cross-dataset evaluation on COCO and Objects365. $^*$ denotes fully-supervised with annotations of all the categories in the target dataset. $\S$ denotes trained only with base classes of LVIS. \jyyoo{$\dag$ denotes reproduced by our implementation, while all the other comparison numbers are reproduced based on their official implementations. For fair comparison, all the methods employed ResNet50 backbone.}% $\dag$: results reported from \cite{rkd}.
%$\ddag$: additionally using pseudo box-level labels w.r.t. novel classes.
}
% \vspace{-1mm}
\vspace{-4mm}
\label{table:table4_transfer}
\end{table}

\subsection{Cross-Dataset Evaluation}
\label{exp:transfer} 
% To validate the generalization capability of expanding detection vocabulary on unseen images, we evaluate the transfer performance, evaluating the models trained with LVIS-base on the other detection datasets (COCO, Object365, and OpenImages). 
To further investigate the open-vocabulary property, we evaluate the transfer performance of our model on the other detection datasets in Table \ref{table:table4_transfer}.
% beyond the evaluation protocol on data splits within the same dataset. %the in-domain evaluation protocol.
\jhaaai{Simply adding our proposed method (the last row)} on the \hskim{LVIS-supervised} detector (2nd row) achieved large performance improvement on COCO (+$1.6\ \text{AP}_{50}$). %, OpenImages ($+2.3\ \text{AP}_{50}$). 
Notably, our method clearly achieves the best transfer performance among the comparisons on COCO, \jhaaai{trailing right behind the \cam{target-supervised case} (1st row) that provides the \cam{upper-bound} performance.}
% On Objects365 and OpenImages, our method shows one of the most superior performance in terms of $AP_{50}$. 
% \jyyoo{We also evaluate $AP_{rare}$ values with $\frac{1}{3}$ classes with the fewest images in each dataset as rare classes along LVIS, as in \cite{detic}.}
\jhaaai{When transferred on Objects365, we achieve similar performance with the state-of-the-art method, F-VLM, while ours clearly attained the top \cam{AP} performance with regard to the rare classes, $\text{AP}^{\text{rare}}_{50}$.}
This result indicates that our model generalizes well even for the rare classes from the different data distributions.
% \jyyoo{This indicates that our model generalizes well when the data distribution is different even for rare classes that are not in the base classes of LVIS with high probability or have very few annotations.}
% the overall performance difference between comparison methods is marginal. Nevertheless, our method exhibits one of the most superior performance among the comparisons.
% since Objects365 contains a lot of annotated objects per image compared to the other datasets, RPN has difficulty in localizing all the objects which leads to low recall rate and AP. Consequently, 
% Furthermore, our method exhibits one of the most superior performance among the comparisons.

% For the case where IN-L is additionally used for training, transfer performances are further improved to trail behind the \hskim{Target}-Supervised case (1st row) that provides the upper bound performance. In particular, our proposed method consistently shows better performance compared to Detic (2nd last row) on all the datasets.
% Especially, our proposed method shows a competitive or even better performance compared to Object-Centric (2nd last entry), which additionally leverages the external model~\cite{mvit} to obtain box annotation for image-level supervision \ctv{where the model is trained on massive amount of classes and hence might overlap with the classes defined within the evaluation datasets.}
% objects365의 경우 한 이미지당 instance 수나 category수가 다른 데이터셋에 비해 많음. 훨씬 dense하여, full-supervised 성능 자체가 낮은 데이터셋. 성능 낮은 이유를 rpn의 recall 자체가 낮기 때문이라고 objects365에서 이유를 꼽음. 아마 애초에 rpn recall이 낮기 때문에 성능 향상이 미미한것으로 보임.

\begin{table}[t]

\centering
\smallskip\noindent
\tabcolsep=0.08cm
\resizebox{1\linewidth}{!}{
% \begin{tabular}{ccc|cc}
% % with AP
% \toprule
% Proxy-novel & Granularity & Prototype & $\text{AP}_r$ & \textcolor{gray}{$\text{AP}$} \\
% \midrule
% \textcolor{red2}{\ding{56}} & - & - & 24.6 & \textcolor{gray}{32.4} \\
% \textcolor{darkergreen}{\ding{52}} & Instance-wise & - & 24.8 & \textcolor{gray}{32.4} \\
% \textcolor{darkergreen}{\ding{52}} & Class-wise & centroid & 25.6 & \textcolor{gray}{32.6} \\
% \textcolor{darkergreen}{\ding{52}} & Class-wise & weighted by IoU & 26.0 & \textcolor{gray}{32.4} \\
% \textcolor{darkergreen}{\ding{52}} & Class-wise & weighted by obj. score & 26.2 & \textcolor{gray}{32.5} \\

% % without AP
\begin{tabular}{ccc|c}
\toprule

\multirow{2}{*}{Proxy-novel} & \multirow{2}{*}{\begin{tabular}[c]{@{}c@{}} Mixup \\ granularity \end{tabular}} & \multirow{2}{*}{\begin{tabular}[c]{@{}c@{}} Visual \\ prototype \end{tabular}} & \multirow{2}{*}{$\text{AP}_r$} \\
& & & \\

% Proxy-novel & Granularity & Prototype & $\text{AP}_r$  \\
\midrule
\textcolor{black}{\ding{56}} & - & - & 24.6 (+0.0)  \\
\textcolor{black}{\ding{52}} & Instance-wise & - & 24.8 (+0.2) \\
\textcolor{black}{\ding{52}} & Class-wise & centroid & 25.6 (+1.0) \\
\textcolor{black}{\ding{52}} & Class-wise & weighted by IoU & 26.0 (+1.4) \\
\textcolor{black}{\ding{52}} & Class-wise & weighted by obj. score & 26.2 (+1.6) \\

\bottomrule

% \begin{tabular}{cc|cc}
% \toprule
% Class-wise Mixup? & Prototype & $AP_r$ & $AP$ \\
% \midrule
% \textcolor{red2}{\ding{56}} & - & 24.6 & 32.4 \\
% \textcolor{darkergreen}{\ding{52}} & \textcolor{red2}{\ding{56}} & 24.8 & 32.4 \\
% \textcolor{darkergreen}{\ding{52}} & Average & 25.6 & 32.6 \\
% \textcolor{darkergreen}{\ding{52}} & IoU & 26.0 & 32.4 \\
% \textcolor{darkergreen}{\ding{52}} & Objectness score & 26.2 & 32.5 \\
% \bottomrule

\end{tabular}
}
\vspace{-0.5mm}
\caption{Effects of mixup granularity and weighting methods to construct \jhaaai{visual} prototype embeddings. We sampled the mixing coefficient $\lambda$ from $\text{Beta}(1, 1)$, when synthesizing proxy-novel classes.}
\vspace{-2.5mm}
\label{table:abs_prototype}
\end{table}

\subsection{\cam{Ablation Studies}}
\label{exp:ablation}

 We perform ablation studies to figure out the efficacy of each component and sensitivity of the hyper-parameter comprising our proposed method, ProxyDet. 
 Unless specified, we use both LVIS and IN-L as training datasets as in~\citet{detic} and report mask \jhaaai{$\text{AP}_{\text{r}}$ on LVIS dataset}. %\jhaaai{We also visualize the open-vocabulary detection results of ProxyDet.}

\noindent{\textbf{Effect of class-wise mixup with prototype construction.}} In Table \ref{table:abs_prototype}, we analyze the effect of proxy loss via class-wise mixup and the technique for constructing the \jhaaai{visual} prototype.
Using instance-wise mixup \cam{(2nd row)} without the construction of class-wise \jhaaai{visual} prototype embeddings shows a marginal performance difference with the vanilla case (1st row).
\cam{Since instance-wise mixup blends individual instances of region embeddings for each base class, it comes with the drawback of incorporating the noisy embeddings from low-quality box proposals (Figure \ref{fig:prototype}) into the mixup without any form of filtering, which might impede stable training.}
\cam{In contrast, leveraging the class-wise prototype embeddings (3-5th rows) largely enhances the performance of novel classes}. Especially, robustly designing the prototype embedding based on IoU or objectness score further strengthens novel class performance by reducing the influence of \cam{such} low-quality region embeddings on prototype construction.

%%%%%%%%%%%%%%%%%%%%%%% EXPERIMENTS 내 approximation %%%%%%%%%%%%%%%%%%%%%%%
% \noindent{\textbf{Towards better approximation of novel classes.}} \ctv{In Table \ref{table:abs_mixup_pairs}, we analyze the effect of selecting the proxy-novel classes that are analogous to the novel classes as presented in Section~\ref{sec:towards_better_novel}. Notably, randomly selecting the proxy-novel classes (2nd entry) showed largely improved performance (+3.7 Non-ovl. AP$_r$) with respect to the truly novel classes (i.e., non-overlapping with IN-L), compared to the vanilla case where proxy loss is not used (1st row). Meanwhile, applying proxy loss only on the proxy-novel classes that are similar with novel classes (last entry) further largely improved by 5.0 Non-ovl. AP$_r$, providing the upper bound performance of our proposed method.}
%%%%%%%%%%%%%%%%%%%%%%% EXPERIMENTS 내 approximation %%%%%%%%%%%%%%%%%%%%%%%

\begin{table}[t]

\centering
\smallskip\noindent
\tabcolsep=0.30cm
\resizebox{1\linewidth}{!}{
% \begin{tabular}{cc|cc}
% \toprule

% Class-wise Mixup? & Sampling Distribution of $\lambda$ & $AP_r$ & $AP$ \\
% \midrule
%  \textcolor{red2}{\ding{56}} & - & 24.6 & 32.4 \\
%  \textcolor{darkergreen}{\ding{52}} & $Bernoulli(\frac{1}{2})$ & 24.8 & 32.6 \\
% \textcolor{darkergreen}{\ding{52}} & $Beta(1, 1)$ & 26.2 & 32.5 \\
% \textcolor{darkergreen}{\ding{52}} & $Beta(5, 5)$ & 26.0 & 32.5 \\
% \textcolor{darkergreen}{\ding{52}} & $Beta(10, 10)$ & 25.5 & 32.4 \\
% \textcolor{darkergreen}{\ding{52}} & $Beta(50, 50)$ & 25.6 & 32.3 \\
% \bottomrule

% % with AP
% \begin{tabular}{ccc|cc}
% \toprule

% Class-wise Mixup & Prototype & Sampling Distribution of $\lambda$ & $\text{AP}_r$ & \textcolor{gray}{$\text{AP}$} \\
% \midrule
%  \textcolor{red2}{\ding{56}} & - & - & 24.6 & \textcolor{gray}{32.4} \\
%  \textcolor{red2}{\ding{56}} & Obj. Score & $\text{Bernoulli}(0.5)$ & 24.8 & \textcolor{gray}{32.6} \\
% \textcolor{darkergreen}{\ding{52}} & Obj. Score & $\text{Beta}(1, 1)$ & 26.2 & \textcolor{gray}{32.5} \\
% \textcolor{darkergreen}{\ding{52}} & Obj. Score & $\text{Beta}(5, 5)$ & 26.0 & \textcolor{gray}{32.5} \\
% \textcolor{darkergreen}{\ding{52}} & Obj. Score & $\text{Beta}(10, 10)$ & 25.5 & \textcolor{gray}{32.4} \\
% \textcolor{darkergreen}{\ding{52}} & Obj. Score & $\text{Beta}(50, 50)$ & 25.6 & \textcolor{gray}{32.3} \\

% without AP
\begin{tabular}{ccc|c}
\toprule

\multirow{2}{*}{\begin{tabular}[c]{@{}c@{}} Class-wise \\ Mixup \end{tabular}} & \multirow{2}{*}{\begin{tabular}[c]{@{}c@{}} Visual \\ prototype \end{tabular}} & \multirow{2}{*}{\begin{tabular}[c]{@{}c@{}} $\lambda$ sampling \end{tabular}} & \multirow{2}{*}{\begin{tabular}[c]{@{}c@{}} $\text{AP}_r$ \end{tabular}} \\
& & & \\

% Class-wise Mixup & Prototype & Sampling Distribution of $\lambda$ & $\text{AP}_r$ \\
\midrule
 \textcolor{black}{\ding{56}} & - & - & 24.6 (+0.0) \\
 \textcolor{black}{\ding{56}} & obj. score & $\text{Bernoulli}(0.5)$ & 24.8 (+0.2) \\
\textcolor{black}{\ding{52}} & obj. score & $\text{Beta}(1, 1)$ & 26.2 (+1.6)\\
\textcolor{black}{\ding{52}} & obj. score & $\text{Beta}(5, 5)$ & 26.0 (+1.4) \\
\textcolor{black}{\ding{52}} & obj. score & $\text{Beta}(10, 10)$ & 25.5 (+0.9) \\
\textcolor{black}{\ding{52}} & obj. score & $\text{Beta}(50, 50)$ & 25.6 (+1.0) \\
\bottomrule

\end{tabular}
}
\caption{Effect of sampling strategy for \jhaaai{mix} coefficient $\lambda$.
% $Bernoulli(\frac{1}{2})$: w/o class-wise mixing. sampling lambda from random uniform was the best. class-wise mixing is essential for novel class performance. Without it, shows same performance with vanilla.
}
% \vspace{-1mm}
\label{table:abs_lambda_study}
\vspace{-4mm}
\end{table}
% \input{tables/abs_mixup_pairs}

% \vspace{-5.5mm}

 \noindent{\textbf{Ablation on mixing coefficient.}} In Table \ref{table:abs_lambda_study}, we analyze the effect of sampling strategy for the coefficient of mixing \jhaaai{base} classes, $\lambda$. Only leveraging \jhaaai{robust visual} prototype without class-wise mixing (2nd row) by sampling $\lambda$ from \hskim{\{0, 1\} following Bernoulli($0.5$)} shows similar performance with the vanilla case (1st row). 
 % However, when class-wise mixing is further utilized by sampling from Beta distribution (3-6th rows) the performance is largely enhanced, where random-uniformly sampling from 0-1 (3rd row) was simply the best performer.
 \jyyoo{However, employing class-wise mixing via Beta distribution sampling (3-6th rows) substantially enhances the performance of novel classes, implying the significance of constructing and utilizing proxy-novel classes through the fusion of base classes to expand the proximity \cam{to the embedding space near} novel classes.}
 % 의미 직접적으로 강조해주면 좋을 것 같아요.
 %--------------------------------------------------------------------------
\begin{figure}[t]
    \centering
    \begin{subfigure}[b]{0.23\textwidth}
        \centering
        \includegraphics[width=\textwidth]{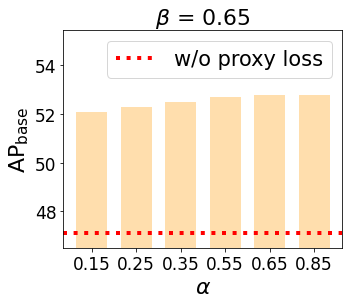}
        % \caption{Similarity distribution}
        \label{fig:abs_alpha_study}
    \end{subfigure}
    % \vspace{-5mm}
    \begin{subfigure}[b]{0.23\textwidth}
        \centering
        \includegraphics[width=\textwidth]{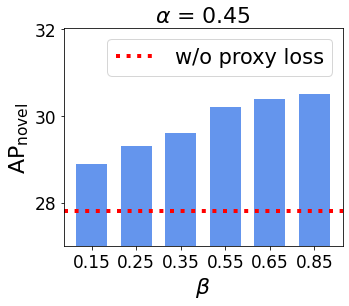}
        % \caption{Similarity distribution}
        \label{fig:abs_beta_study}
    \end{subfigure}
    \vspace{-6mm}
    \caption{Ablation on fusion parameters $\alpha$ and $\beta$. We report AP results on Open-Vocabulary COCO benchmark.
    % Larger the $\alpha$, $\beta$, larger performance improvement. conducted on OV-COCO based on Detic.
    }
    \label{fig:abs_fusion_study}
    \vspace{-2mm}
\end{figure}
%--------------------------------------------------------------------------
 \begin{table}[t]

\centering
\smallskip\noindent
\tabcolsep=0.35cm
% \vspace{-1mm}
\resizebox{0.85\linewidth}{!}{

% % with AP
% \begin{tabular}{cc|cc}
% \toprule
% LVIS Mixup & IN-L Mixup & $\text{AP}_r$ & \textcolor{gray}{$\text{AP}$} \\
% \midrule

% \textcolor{darkergreen}{\ding{52}} & \textcolor{red2}{\ding{56}} & 26.2 & \textcolor{gray}{32.5} \\
% \textcolor{darkergreen}{\ding{52}} & \textcolor{darkergreen}{\ding{52}} & 23.6 & \textcolor{gray}{31.9} \\

% % without AP
\begin{tabular}{cc|c}
\toprule
LVIS Mixup & IN-L Mixup & $\text{AP}_r$ \\
\midrule

\textcolor{black}{\ding{52}} & \textcolor{black}{\ding{56}} &  \ 26.2 (+0.0) \\
\textcolor{black}{\ding{52}} & \textcolor{black}{\ding{52}} & 23.6 (-2.6) \\
\bottomrule

\end{tabular}
}
\vspace{-1mm}
\caption{Effect of class-wise mixup on IN-L.
%reason of degeneration: max-size proposal is too coarse. (multi-objects, etc) & not enough proposals for forming prototype embedding. (one max-size proposal)
}
% \vspace{-1mm}
\label{table:abs_IN-L_mixup}
\vspace{-4mm}
\end{table}

% \vspace{-4mm}
% \input{tables/abs_mixup_pairs}
\noindent{\textbf{Effect of score fusion parameters.}} In Figure \ref{fig:abs_fusion_study}, we analyze the effect of score fusion parameters applied with regard to the detector head trained with our proxy loss (Eq \ref{eq:score_fusion}). Generally, all different $\alpha$ or $\beta$ cases stably outperform the vanilla case (red dotted line) where the detector head trained with our proxy loss does not participate in score fusion (i.e., $\alpha=0, \beta=0$).
Moreover, as the value of $\alpha$ is raised, the performance of base classes 
 increases gradually while raising $\beta$ results in a more rapid improvement in terms of the performance for novel classes. This indicates the essential role of training with our proxy loss in terms of improving performance for both base and, more significantly, novel classes.
% Also, as $\alpha$ or $\beta$ increases, base or novel class performance increases gradually. This indicates the essential role of training with our proxy loss in terms of improving performance for both novel and base categories.
 
 % the final classification score is more weighted on outputs from the head trained with proxy loss for base and novel classes

 \noindent{\textbf{\cam{Scale-up} of mixup on ImageNet-LVIS.}} In Table \ref{table:abs_IN-L_mixup}, we validate the effect of additionally applying class-wise mixup on IN-L~\cite{detic}. \jhaaai{Since IN-L does not provide the human-labeled bounding box annotations}, we simply leveraged max-size proposal from RPN following \citet{detic} to construct the \jhaaai{visual} prototype embedding for our proxy loss. We observed that performance of the novel classes largely degrades. We conjecture that since max-size proposal is too coarsely localized, it might contain the other class objects and non-objects which can be served as noise to construct the visual prototype embedding for proxy loss.

 \noindent{\textbf{Comparison with the other supervision \jhaaai{for novel classes}.}}
 In Table \ref{table:abs_distill_comparison}, we compared \jyyoo{the} performance with the other supervision \jhaaai{technique} which distills the knowledge \jhaaai{of novel class objects} from the outputs of CLIP visual encoder~\cite{vild}.
 % Applying BCE loss with respect to the base classes (2nd entry) only showed marginal difference with the vanilla case (1st entry) which does not employ the separated detector head.
 By applying distillation supervision along with the vanilla BCE loss, novel class performance 
 %improves 
 \jyyoo{is improved} by +$0.7$ AP$_r$. Meanwhile, adding our proposed proxy loss shows the best performance, further improving by +$1.6\ \text{AP}_r$.

 \begin{table}[t]

\centering
\smallskip\noindent
\tabcolsep=0.35cm

\vspace{1.0mm}
\resizebox{0.95\linewidth}{!}{

% % with AP
% \begin{tabular}{l|cc}

% \toprule

% \multicolumn{1}{c|}{Supervision} & $\text{AP}_r$ & $\text{AP}$ \\
% \midrule

% BCE loss (base classes) & 24.6 & \textcolor{gray}{32.4} \\
% % BCE loss (base classes)  & 24.5 &  32.5  \\
% BCE + Distillation~\cite{vild} & 25.3 & \textcolor{gray}{32.2} \\
% BCE + Proxy loss (ours, Eq.~\ref{eq:proxy_loss}) & 26.2 & \textcolor{gray}{32.5} \\

% % without AP
\begin{tabular}{l|c}

\toprule

\multicolumn{1}{c|}{Supervision} & $\text{AP}_r$ \\
\midrule

BCE loss (base classes) & 24.6 (+0.0) \\
% BCE loss (base classes)  & 24.5 &  32.5  \\
BCE + distillation~\cite{vild} & 25.3 (+0.7) \\
BCE + proxy loss (ours, Eq.~\ref{eq:proxy_loss}) & 26.2 (+1.6) \\
\bottomrule

\end{tabular}
}
\caption{Comparison of proxy-novel supervision with knowledge distillation from CLIP.}
\label{table:abs_distill_comparison}
\vspace{-1mm}
\end{table}

% \noindent{\textbf{Qualitative Results.}} In the Appendix, we present visualization of detection results for our detector, ProxyDet.
% % We trained on LVIS dataset and transferred to the other datasets (Objects365 and OpenImages) without fine-tuning.
% ProxyDet is capable of localizing and classifying base and novel categories. Especially for the novel categories, ProxyDet is able to expand the detection vocabulary into the fine-grained novel classes, e.g., rice cooker, traffic cone, and gas stove.

\subsection{Towards Better Approximation of Novel Class}
\label{sec:towards_better_novel}
In this subsection, we take a deeper look at the effectiveness of proxy-novel by assuming that \jhaaai{a set of vocabularies including} the names of the novel categories are given as prior knowledge.
With the knowledge of the novel classes and their corresponding text embeddings, we can control which base classes \jhaaai{should be mixed to more accurately approximate each of the target novel \cam{classes}.}
% create \hskim{embeddings} of the proxy-novel, bringing them closer to the target novel classes. 
%With the knowledge of the novel classes and their corresponding textual embeddings, we can only leverage the ideal proxy novel classes that are closer to the \hskim{target} novel classes.
\jhaaai{Specifically}, for each novel class, we extract the \jhaaai{optimal} combination of base classes that exhibit the highest similarity to the novel class \jhaaai{in CLIP textual embedding space}. Subsequently, we employ the identified combination of base classes to synthesize the proxy-novel and implement our proposed proxy loss.

\jhaaai{As shown in Table \ref{table:abs_mixup_pairs}, selecting the optimal combination of base classes nearest to each of novel \cam{classes} improves AP$_r$ by +2.3 compared to the vanilla case (1st row), achieving better performance compared to the random selection strategy (2nd row). However, in real-world scenarios of OVOD, the knowledge about the names of novel categories is not given during \cam{the} training phase and hence the random selection is a more practical strategy for generating proxy-novel classes. % Figure 2에서 보여준거처럼 잘 approximate하니까.
Under this real-world scenario without explicitly knowing the novel categories, enlarging the set of prior vocabularies \cam{for mixing} to \cam{ideally} cover all the novel categories is a practical alternative, which is our future work.}
% Ideally, 
% To deal with this, as a future work, enlarging the set of prior vocabularies without explicitly knowing the novel classes ideally cover all the novel classes  contain the 
% we mainly use }
% The comparison of effectiveness between controlled proxy-novel and randomly mixed proxy-novel is presented in Table~\ref{table:abs_mixup_pairs}.
% We observe that despite a marginal performance difference of only 0.7 $\text{AP}_r$ between the two mixing strategies, there is a significant improvement in performance when using our proxy-novel approach for training (e.g., +1.6 $AP_r$ with random selection). Although we mainly use the randomly mixing strategy for generating proxy-novel classes due to the lack of knowledge about the names of novel categories in real-world scenarios of OVOD, the results indicate that our framework sufficiently allows the detector to classify novel classes that are not present in the training.

\begin{table}[t]

\centering
\smallskip\noindent
\tabcolsep=0.08cm
\resizebox{0.95\linewidth}{!}{
\begin{NiceTabular}{ccc|c}
\toprule
\multirow{2}{*}{\begin{tabular}[c]{@{}c@{}} Proxy-novel \end{tabular}} & \multirow{2}{*}{\begin{tabular}[c]{@{}c@{}} \jhaaai{Novel category} \\ \jhaaai{prior} \end{tabular}} & \multirow{2}{*}{\begin{tabular}[c]{@{}c@{}} \jhaaai{Mixing class} \\ \jhaaai{selection} \end{tabular}} & 
% \multicolumn{3}{c}{$\text{AP}_r$}\\
% \cmidrule(lr){4-6}
\multirow{2}{*}{\begin{tabular}[c]{@{}c@{}} AP$_r$ \end{tabular}}  \\
& & &  \\
\midrule
\textcolor{black}{\ding{56}} & \textcolor{black}{\ding{56}} & - & 24.6 (+0.0)  \\ 
\textcolor{black}{\ding{52}} & \textcolor{black}{\ding{56}} & random & 26.2 (+1.6)  \\ 
\textcolor{black}{\ding{52}} & \textcolor{black}{\ding{52}} & novel-nearest & 26.9 (+2.3) \\ 
%%%%%%%%%%%% AP_r without overalpped or not. %%%%%%%%%%%%

% \textcolor{red2}{\ding{56}} & base nearest & & & 25.4 \\
% \textcolor{red2}{\ding{56}} & random & 27.0 & 23.6 & 26.3 \\
% \midrule
% \textcolor{darkergreen}{\ding{52}} & novel nearest & 27.0 & 25.4 & 26.7 \\
\bottomrule
\end{NiceTabular}

% \begin{tabular}{cc|ccc}
% \toprule
% Novel category? & Mixup class 
%  pair & Ovl $\text{AP}_r$ & Non-ovl $\text{AP}_r$ & Overall $\text{AP}_r$ \\
% \midrule
% \textcolor{red2}{\ding{56}} & base farthest & 25.6 & 13.9 & 23.4 \\ 
% \textcolor{red2}{\ding{56}} & base nearest & & & 25.4 \\
% \textcolor{red2}{\ding{56}} & random & 27.0 & 23.6 & 26.3 \\
% \midrule
% \textcolor{darkergreen}{\ding{52}} & novel nearest & 27.0 & 25.4 & 26.7 \\
% \bottomrule
% \end{tabular}
}
\vspace{-0.8mm}
\caption{The effect of \jhaaai{selection strategy for the base classes to be mixed}.
% \vspace{0.9mm}
% Used IN-L from Detic~\cite{detic}. Ovl $AP_r$: AP w.r.t. novel classes in the IN-L. novel nearest: requires novel category.
}
\vspace{-2mm}
\label{table:abs_mixup_pairs}
\end{table}

\section{Conclusion}

  Motivated by the observation that novel class representations can be approximated through mixup between base classes, we devise a new learning strategy to effectively \cam{detect novel category objects.}
  % \hskim{\sout{Through the construction of robust prototype embeddings prototype for base classes and the class-wise mixup strategy for synthesizing proxy novel classes, we can approximately learn on the novel class distribution.}}
  Our method introduces proxy loss on proxy-novel classes synthesized by well-defined prototypes of base classes, which encourages our models to explore the embedding space close to the embeddings of novel classes. 
  By the extensive experiments on various benchmarks and rigorous analysis with ablations, we demonstrate that our simple add-on technique helps generalization for \hsjeong{detecting} broad range of novel classes. % We hope our work facilitates a new perspective for OVOD task along with leveraging pseudo-labeling on the massive amount of data.

% \begin{itemize}
%     \item \todo{ I motivate the problem and discuss existing solutions limitations again, but in a very short paragraph. }
%     \item \todo{ I summarize what I am proposing in the paper and discuss what I discovered from experiments results, difficulties and even discuss drawbacks of my approach. Additionally, I let it clear what are the contributions my paper is giving to new researchers of the area. }
%     \item \todo{ I discuss about future work that can be inspired from my research reported in the paper, such as modifying other parameters, trying different classifiers, making the datasets more difficult and even studying the possibility of transferring my solution to another problem. }
% \end{itemize}

% for arxiv only
\section{Additional Analysis}
\noindent\textbf{{Number of the pseudo-labeled novel classes.}}
In Table \ref{table:overlap_analysis}, we analyze the ratio of LVIS-novel classes that are pseudo-labeled on IN-L dataset. This supplementary data involves a significant amount of image-level labels for novel classes, about 82\% of the total novel classes.

\begin{table}[h]

\centering
\smallskip\noindent
\tabcolsep=0.1cm
\caption{\textbf{Number of novel classes divided by whether pseudo-labeled or not on IN-L dataset.}
}
\resizebox{0.55\linewidth}{!}{
\begin{NiceTabular}{ccc}
\toprule
\multicolumn{3}{c}{LVIS-novel classes}\\
\midrule
% \cmidrule(lr){1-3}
Non-pseudo & Pseudo & Overall  \\
        
\toprule

\multirow{2}{*}{\begin{tabular}[c]{@{}c@{}} 60 \\ (18\%) \end{tabular}} &
\multirow{2}{*}{\begin{tabular}[c]{@{}c@{}} 277 \\ (82\%) \end{tabular}} &
\multirow{2}{*}{\begin{tabular}[c]{@{}c@{}} 337 \\ (100\%)  \end{tabular}} \\ 
& &   \\

\bottomrule

\end{NiceTabular}
}
\label{table:overlap_analysis}
\end{table}

\noindent\textbf{{Effect of the other pseudo-labeling method.}} In Table \ref{table:rkd_table1_nonovl_ap}, we analyze performance of another pseudo-labeling method~\cite{rkd} in terms of novel classes depending on whether pseudo-labeled or not on the supplementary dataset they employed, IN-L. Although the performance of pseudo-labeled novel classes is largely improved by $6.9\%$, the performance of non-pseudo labeled novel classes only considerably degrades by $2.6\%$. The result is consistent with that of Table 1 in the manuscript, which indicates that the pseudo-labeling methods still lack generalization on the novel classes that are completely unseen during training.
% 2. table 1을 VLDet, RKD에 대해서 돌린 결과 (non-ovl에 대한 성능 향상 marginal 여부) @김희수
% RKD의 baseline, baseline + pseudo-labeling 성능 비교
% 3. LVIS Novel <=> IN-L overlap 비율 비교. TBU @김희수

\begin{table}[h]

\centering
\smallskip\noindent
\tabcolsep=0.1cm
\caption{\textbf{Effectiveness of pseudo-labeling on the generalization performance towards novel classes.} For evaluation, we adopt the same setting depicted on Table 1 of the manuscript.
}
\resizebox{0.95\linewidth}{!}{
\begin{NiceTabular}{c|ccc}
\toprule
\multirow{2}{*}{\begin{tabular}[c]{@{}c@{}} Method \end{tabular}} & 
\multicolumn{3}{c}{$\text{AP}_r$}\\
\cmidrule(lr){2-4}
& Non-pseudo & \textcolor{gray}{Pseudo} & Overall  \\
        
\midrule
% 11.621
% 18.572
% 17.244

% 14.230
% 11.672
% 12.161

\multirow{2}{*}{\begin{tabular}[c]{@{}c@{}}  w/o pseudo-labeling~\cite{rkd}  \end{tabular}} & 
\multirow{2}{*}{\begin{tabular}[c]{@{}c@{}} 14.2 \\ (+0.0) \end{tabular}} &
\multirow{2}{*}{\begin{tabular}[c]{@{}c@{}} \textcolor{gray}{11.7} \\ \textcolor{gray}{(+0.0)}
\end{tabular}} &
\multirow{2}{*}{\begin{tabular}[c]{@{}c@{}} \textcolor{black}{12.2} \\ (+0.0) \end{tabular}} \\ 
& & &   \\

\multirow{2}{*}{\begin{tabular}[c]{@{}c@{}} w/ pseudo-labeling~\cite{rkd}  \end{tabular}} & 
\multirow{2}{*}{\begin{tabular}[c]{@{}c@{}} 11.6 \\ \textbf{\textcolor{red2}{(-2.6)}} \end{tabular}} &
\multirow{2}{*}{\begin{tabular}[c]{@{}c@{}} \textcolor{gray}{18.6} \\ \textcolor{gray}{(+6.9)}
\end{tabular}} &
\multirow{2}{*}{\begin{tabular}[c]{@{}c@{}} 17.2 \\ \textcolor{darkergreen}{(+5.0)} \end{tabular}} \\ 
& & &   \\
\bottomrule

\end{NiceTabular}
}
% \vspace{-3mm}
\label{table:rkd_table1_nonovl_ap}
\end{table}

\begin{figure}[h]
    \centering
        \centering
        \includegraphics[width=0.4\textwidth]{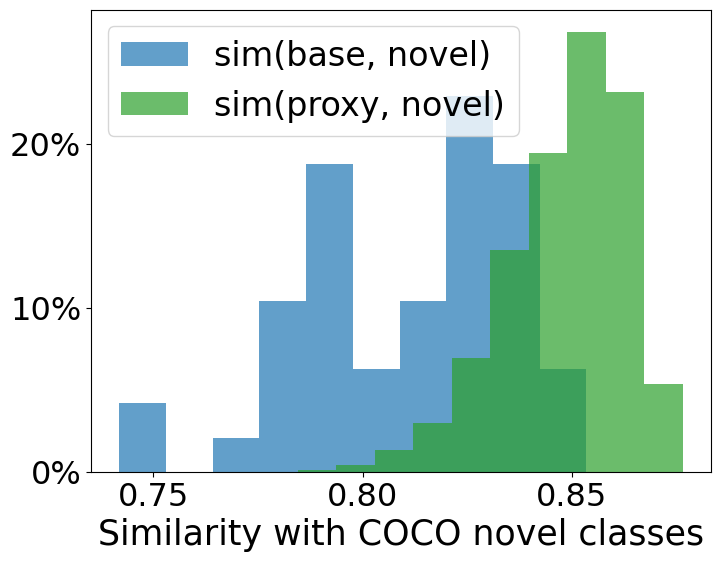}
        \caption{\textbf{Histogram of similarity between training class groups (base and proxy-novel) and novel classes in OV-COCO dataset.} For visualization, we adopt the same setting depicted in Figure 2 of the manuscript.}
        % For each novel class, we compute top-1 cosine similarity with each category group, using text representations obtained from CLIP~\cite{clip}. The x-axis is sorted by ascending order of similarity score between base and novel classes.
        \label{fig:novel_similarity_distribution_coco}
    \label{fig:novel_similarity_coco}
\end{figure}

\noindent\textbf{{Similarity of proxy-novel to novel classes on OV-COCO.}}
In Figure \ref{fig:novel_similarity_coco}, we additionally analyze  similarity of proxy-novel classes to the novel classes on OV-COCO dataset. Compared to the existing base classes, proxy-novel classes exhibit higher similarity with novel classes, which is consistent with the tendency on OV-LVIS dataset (Figure 2 of the manuscript). Through additional training with these proxy-novel classes, our model can effectively explore the proximal representation space of novel classes.

% 1. coco에서의 similarity 결과.
    % histogram of figure 2는 X. rebuttal떄 필요시 사용하는것으로.

\begin{figure*}[t]
    \centering
    \includegraphics[width=1\linewidth]{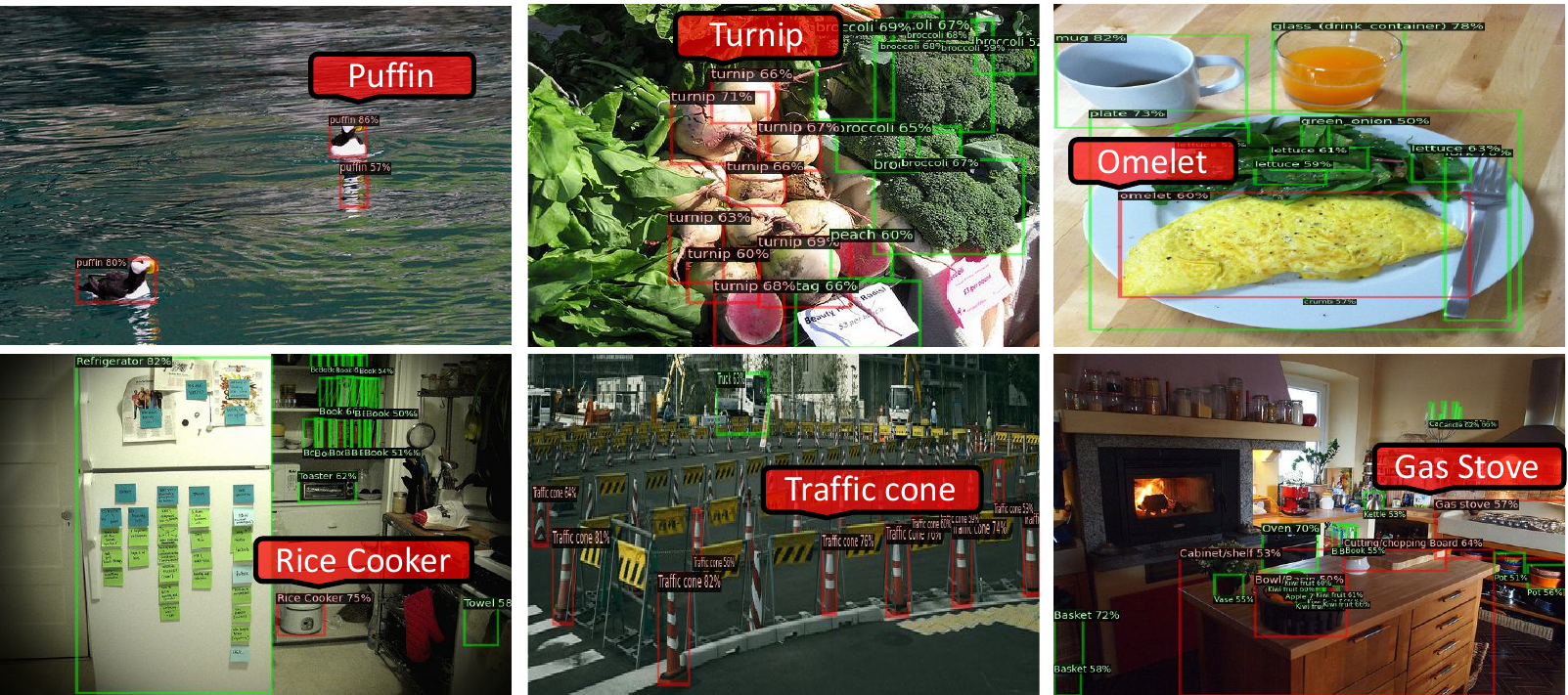}
    % \vspace{2mm}
    \caption{\textbf{Qualitative results on LVIS (top) and Objects365 (bottom) datasets.} For clarity, we divide detection results by green color (base classes) and red color (novel classes).
    % Left: detection results on LVIS dataset.
    %Center: transfer detection results on Objects365. Right: transfer detection results on OpenImages.
    }
    \label{fig:qualitative_results}
    \vspace{2mm}
\end{figure*}

\noindent\textbf{{Visualization of detection results.}} In Figure \ref{fig:qualitative_results}, we visualize the open-vocabulary detection results of our proposed detector, ProxyDet. Notably, while our ProxyDet can localize all the base classes seen in the training phase, it can also capture the fine-grained novel objects including \cam{rice cooker, traffic cone, and gas stove}.

% \noindent{\textbf{Qualitative Results.}} In the Appendix, we present visualization of detection results for our detector, ProxyDet.
% % We trained on LVIS dataset and transferred to the other datasets (Objects365 and OpenImages) without fine-tuning.
% ProxyDet is capable of localizing and classifying base and novel categories. Especially for the novel categories, ProxyDet is able to expand the detection vocabulary into the fine-grained novel classes, e.g., rice cooker, traffic cone, and gas stove.

% \noindent\textbf{Qualitative examples of novel class approximations}. In Figure \ref{fig:mixing_sim}, we mixed a series of animal base classes (\texttt{cat}, \texttt{lion}) using CLIP text embedding, to compare similarity with that of a similar novel class (\texttt{leopard}) and a completely dissimilar novel class (\texttt{refrigerator}). The result informs that 
% both of the novel classes can be better approximated via our proposed class-wise mixup scheme (i.e., $0 < \lambda < 1$).
% The outlier class cannot be satisfactorily approximated, while the in-distribution class can be better approximated by class-wise mixup between similar base classes. 

% \begin{figure}[h]
%     \centering
%         \centering
% \includegraphics[width=\linewidth]{figures/mixing_sim.pdf}
%         \caption{\textbf{Similarity plot between mixed classes and target novel classes.}}
%     \label{fig:mixing_sim}
% \end{figure}

\section{Additional Experimental Results}
\noindent\textbf{{Ablation study on proxy loss.}}
% 4. proxy loss - L2, cosine similarity maximize 등 실험 결과
In Table \ref{table:abs_loss}, we further conduct an ablation study on the design choice of our proposed proxy loss (Eq. 4 in the manuscript). Firstly, we maximize cosine similarity between the visual and textual embedding of the proxy-novel class:

\begin{equation}
    \mathcal{L}_{\text{proxy-cos}} = 1 - \big(\text{sim}(\mathcal{T}(c^{\text{proxy}}), r^{\text{proxy}})\big)
    % \mathcal{T}(c^{\text{proxy}}) - r^{\text{proxy}} ||_{1}, \\
\label{eq:proxy_loss_cosine}
\end{equation}
where $\text{sim}(a, b) = \frac{a^{\top}b}{||a||||b||}$. For the L2 Distance case, we simply apply L2 distance instead of L1 distance in Eq. 4 of the manuscript. The result shows that L1 distance was the best performer by a large margin in terms of novel classes and hence we adopt L1 distance for our proposed proxy loss.

\begin{table}[t]

\centering
\smallskip\noindent
\tabcolsep=0.1cm

\caption{\textbf{Effect of various designs for the proxy loss.}}
\resizebox{0.6\linewidth}{!}{

\begin{tabular}{l|cc}

\toprule

\multicolumn{1}{c|}{Loss} & $\text{AP}_r$ & \textcolor{gray}{$\text{AP}$} \\
\midrule

Cosine similarity & 25.3 & \textcolor{gray}{32.4} \\
L2 distance & 25.1 & \textcolor{gray}{32.5} \\
L1 distance & 26.2 & \textcolor{gray}{32.5} \\

% BCE loss (base classes) & 24.6 & 32.4 \\
% % BCE loss (base classes)  & 24.5 &  32.5  \\
% BCE + Distillation~\cite{vild} & 25.3 & 32.2 \\
% BCE + Proxy loss (ours, Eq.~\ref{eq:proxy_loss}) & 26.2 & 32.5 \\
\bottomrule

\end{tabular}
}

\label{table:abs_loss}
% \vspace{-2mm}
\end{table}

\section{Implementation Details}
% further implementation details.
% M (dimension of image, text embedding) = 512 for all experiments on LVIS and COCO.
% LVIS alpha, beta 명시해주기
% BCE Loss 설명 추가
% positive proposal IoU 값 설명 
% F-VLM supple참고해서, Detic에서 차용한 세팅 명시 (OV-COCO, OV-LVIS)

In Table \ref{table:hyperparams}, we summarize the hyper-parameters for training and evaluating our proposed framework, ProxyDet. Overall, we adopt the same hyper-parameters from Detic~\cite{detic}, except for our own hyper-parameters ($\alpha, \beta, \lambda$). For IoU threshold to filter out positive proposals in OV-LVIS, we apply different thresholds for each stage in cascade head structure~\cite{cascade_rcnn}. \cam{For the evaluation on Objects365 dataset, we measured box mAP for all the classes ($\text{AP}_{50}$) and rare classes ($\text{AP}^{\text{rare}}_{50}$).}

\begin{table}[h]

\centering
\smallskip\noindent
\tabcolsep=0.1cm

\caption{\textbf{ProxyDet hyper-parameter setup.}}
\resizebox{1\linewidth}{!}{

\begin{tabular}{ccc}

\toprule

\textbf{Setup} & \textbf{OV-COCO} & \textbf{OV-LVIS} \\
\midrule

\cam{Prompt for CLIP text encoder} & \cam{\textit{"a \{class\}"}} & \cam{\textit{"a \{class\}"}} \\
Embedding dim ($M$) & 512 & 512 \\
Cascade head & none & 3 stages \\
Positive proposal IoU threshold & 0.5 & [0.6, 0.7, 0.8] \\
Optimizer & SGD & ADAMW \\
Momentum & 0.9 & 0.9 \\
Weight decay & 1e-4 & 1e-4 \\
Learning rate (LR) & 2e-2 & 2e-4 \\
% LR decay factor & 0.1 & 0.1 \\
% LR decay schedule & [60k, 80k] & [30k] \\
\cam{LR decay schedule} & \cam{MultiStep} & \cam{CosineAnnealing} \\
% Warmup LR / steps &  & \\
Training iters & 90k & 90k \\
% Batch size & 16 & 64 \\
% Augmentation & H-flip & H-flip, LSJ\\
NMS threshold & 0.5 & 0.5 \\
Base score fusion weight $\alpha$ & 0.45 & 0.15 \\
Novel score fusion weight $\beta$ & 0.65 & 0.35 \\
Mixing Coefficient $\lambda$ & Beta(1, 1) & Beta(1, 1) \\
% further implementation details.
% M (dimension of image, text embedding) = 512 for all experiments on LVIS and COCO.
% LVIS alpha, beta 명시해주기
% BCE Loss 설명 추가 -> 굳이 필요없을듯
% positive proposal IoU 값 설명 
% F-VLM supple참고해서, Detic에서 차용한 세팅 명시 (OV-COCO, OV-LVIS)

\bottomrule

\end{tabular}
}

\label{table:hyperparams}
% \vspace{-2mm}
\end{table}

\section{Limitations}
\cam{Practically, the potential range of novel classes is vast, encompassing all the classes not included in the base classes presented during training. 
To cover this wide range of novel classes, one might consider scaling up our mixup strategy by mixing a much broader set of base classes sourced from easily accessible and larger datasets.
In an attempt to do this, we applied our mixup on a more extensive dataset, ImageNet-LVIS, but only to degrade novel class performance due to the lack of precise bounding box annotations (Table 7 of the manuscript).
Consequently, developing an advanced method for generating accurate pseudo-box labels to effectively scale up our mixup scheme would be a promising direction for future research.}

% \noindent\textbf{{Cross dataset evaluation}}
% LVIS to COCO: LVIS-base ($\alpha=0.45, \beta=0.65$), LVIS-base + IN-L ($\alpha=0.45, \beta=0.65$)
% LVIS to Objects365: LVIS-base ($\alpha=0.15, \beta=0.25$), LVIS-base + IN-L ($\alpha=0.15, \beta=0.25$)
% LVIS to OpenImages: LVIS-base ($\alpha=0.75, \beta=0.75$), LVIS-base + IN-L ($\alpha=0.75, \beta=0.75$)
\section*{Acknowledgments}

This work was supported by Institute of Information \& communications Technology Planning \& Evaluation (IITP) grant (No.2019-0-00075, Artificial Intelligence Graduate School Program (KAIST), No.2022-0-00713, Meta-learning applicable to real-world problems) and the National Research Foundation of Korea (NRF) grants (No.RS-2023-00209060, A Study on Optimization and Network Interpretation Method for Large-Scale Machine Learning) funded by the Korea government (MSIT). This work was also supported by KAIST-NAVER Hypercreative AI Center.

\bibliography{aaai24}

\end{document}